\documentclass{llncs}
\usepackage{times}
\usepackage{mathtools} 
\usepackage{changepage}
\usepackage{enumitem} 
\usepackage{xcolor} 
\usepackage{amssymb}    
\usepackage{caption}  
\usepackage{tcolorbox}
\tcbuselibrary{raster,skins}
\newcommand{\inducedR}{\ensuremath{f_\textsf{R}}}
 
\newcommand{\AS}{\textsf{S}} 
\newcommand{\decreaser}{\textsf{dec}}
\newcommand{\increaser}{\textsf{inc}}
\newcommand{\expr}{\textsf{exp}} 
\newcommand{\inducedRp}{\ensuremath{f_\textsf{\Rp}}}

\newcommand{\cst}{\ensuremath{\textsf{cst}}}
\newcommand{\sat}{\textsf{sat}}

\newcommand{\num}{\textsf{num}}
 
\newcommand{\nums}{\textsf{Nums}}

\newcommand{\pve}{\textsf{pos}}
\newcommand{\nve}{\textsf{neg}}

\newcommand{\Ainit}{\ensuremath{A_{\textbf{0}}}} 
\newcommand{\na}{\ensuremath{\textsf{n}^a}}
\newcommand{\nainit}{\ensuremath{\na_{\textbf{0}}}}

\newcommand{\Rp}{\ensuremath{R_{\mathbf{p}}}}

\newcommand{\semsmall}{\textsf{sem}}
\newcommand{\fsem}{\ensuremath{f_{\semsmall}}}

\newcommand{\fe}{\ensuremath{f_{\textbf{E}}}}

\newcommand{\co}{\textsf{co}}
\newcommand{\pr}{\textsf{pr}}
\newcommand{\gr}{\textsf{gr}}

\newcommand{\APA}{\textsf{APA}}

\newcommand{\nr}{\ensuremath{\textsf{n}^{r}}}

\newcommand{\Na}{\ensuremath{N^a}}
\newcommand{\Dung}{\textsf{D}}

\newcommand{\Sem}{\textsf{Sem}}

\newcommand{\hs}{\textsf{hs}}
\newcommand{\hide}[1]{} 
  
\newcommand{\orC}{\textsf{or}}
\newcommand{\andC}{\textsf{and}}

\begin{document}
     \title{Numerical Abstract Persuasion Argumentation
     for Expressing Concurrent 
     Multi-Agent Negotiations}  

\author{Ryuta Arisaka \and Takayuki Ito}
\institute{Nagoya Institute of Technology, Nagoya, Japan\\ 
    \email{ryutaarisaka@gmail.com, ito.takayuki@nitech.ac.jp}
}

\maketitle

%\titlenote{Produces the permission block, and
%  copyright information}
%\subtitle{Extended Abstract}
%\subtitlenote{The full version of the author's guide is available as
%  \texttt{acmart.pdf} document}

%\author{Ryuta Arisaka
%    \institute{ryutaarisaka@gmail.com} 
    %\institute{Universidade Federal do Cear{\'a}, 
    %Brazil, email: samysoares@gmail.com}  
%\institute{Zhejiang University, China, email: baiseliao@zju.edu.cn} 
%\institute{National Institute of Informatics, 
    %Japan, email: ksatoh@gmail.com}
%\and Yi N. Wang %\institute{Zhejiang University, China, email: 
%    %ynw@xixilogic.org} 
%\institute{}
%    }

\begin{abstract}   
   A negotiation process 
   by 2 agents $e_1$ and $e_2$ 
   can be interleaved 
   by another negotiation process 
   between, say, $e_1$ and $e_3$.   
   The interleaving may alter  
   the resource allocation assumed 
   at the inception of the first negotiation process.  
   Existing proposals for argumentation-based negotiations
   have focused primarily on two-agent bilateral negotiations, but 
   scarcely on the concurrency 
   of multi-agent negotiations. 
   To fill the gap, we present 
   a novel argumentation theory, 
   basing its development on  
   abstract persuasion argumentation (which 
   is an abstract argumentation formalism with 
   a dynamic relation). 
   Incorporating into it
   numerical information  
   and a mechanism of handshakes among members of the dynamic 
   relation,
   we show that the extended theory adapts well to 
   concurrent multi-agent negotiations over
   scarce resources. 
\end{abstract} 
\section{Introduction}  
Agent negotiations may be modelled 
by game-theoretical approaches, heuristic-based approaches or 
argumentation-based approaches 
\cite{Rahwan03}. 
For obtaining rational explanations as to why 
agents (have) come to a deal, the last,
argumentation-based methods 
(e.g. \cite{Rahwan03,Amgoud07,Hadidi10,Dung08,Kakas06}), 
with their intrinsic strength
to handle information in conflict, have been 
shown to offer some research perspective and direction. 

However, the focus of argumentation-based
formal approaches has been primarily
on bilateral negotiations involving 
only two parties 
so far: extension to more general 
multiagent negotiations is not automatic 
when demand and supply of scarce resources 
being negotiated over 
can change in the middle of a negotiation 
due to negotiation process interleaving. 
Consider the following with agents 
$e_1$, $e_2$, $e_3$.  
%\vspace{0.1cm} 
\begin{itemize}
    \item[] Of $e_1$:     
         She owns an electronics shop. She has recently 
         obtained two Nintendo Switch. Her ask price 
         for the console is \$300 each. 
         She has \$0. \\
    \item[] Of $e_2$:    
         He and his brother have been a big Nintendo fan. He has heard 
         that it is now available at $e_1$'s store. 
         He plans to buy two of them, one 
         for himself, and the other for his brother. 
         His budget is \$650. He currently 
         does not have a Nintendo Switch.\\
    \item[] Of $e_3$:   
         He is a reseller for profit. He has heard 
         Switch consoles are now at $e_1$'s 
         shop for \$300. He has also heard 
         that $e_2$ is looking for the console. 
         He plans to buy them up from $e_1$ and resells 
         them to $e_2$ for \$400 each. 
         He has \$600, and no Nintendo Switch.  
\end{itemize} 
Suppose the preference 
of each agent is such that:  
\begin{itemize} 
    \item $e_1$ likes to sell both of the Switch consoles. The price per console (\$300) is not negotiable. 
    \item $e_2$ likes to obtain two Switch consoles. 
          In case of multiple offers,
          he chooses the cheapest 
          one. If there is only 
          one seller, his default action 
          is to accept  
          any ask price, except when 
          he cannot afford it, in which case
          he asks the seller to lower the price 
          to the amount he possesses. 
    \item $e_3$ likes to obtain as many 
    Switch consoles from $e_1$ for \$300 as his budget 
    allows him, and likes to resell them 
    to $e_2$ for \$400 so long as 
    $e_2$ accepts the deal. He is willing to 
    negotiate over the price if $e_2$ 
    complains at the ask price. However, 
    he will not lower the price below \$300. 
\end{itemize} 
\begin{center} 
    \includegraphics[scale=0.27]{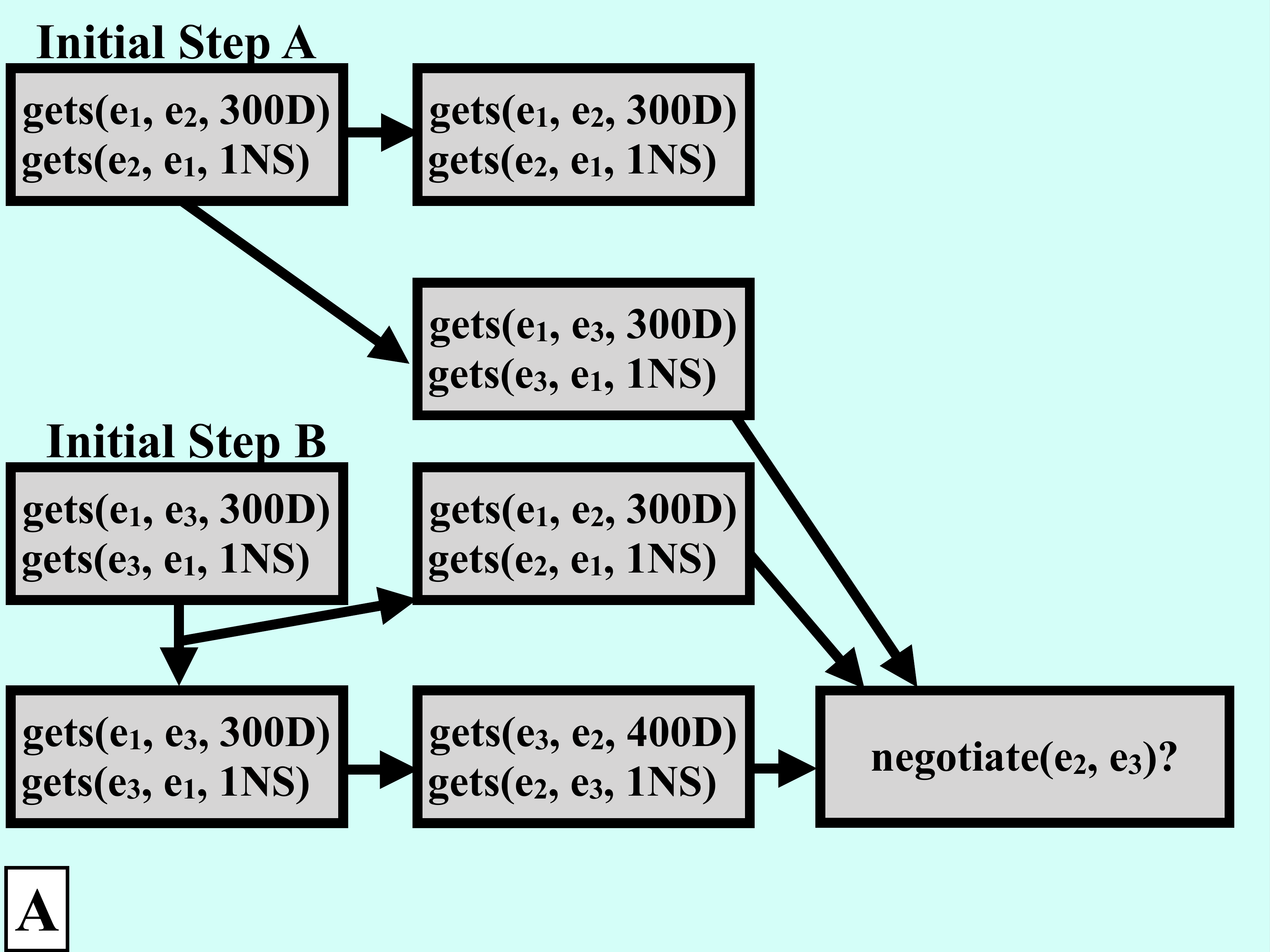}
\end{center} 
Suppose $e_2$ and $e_3$ are the only 
ones who would visit $e_1$ for Switch 
consoles and that both of the consoles 
would be eventually sold to one of them, 
the steps indicated in \fbox{A}  cover 
all possible situations that could result. 
 gets$(e_x, e_y, nD)$ means 
for $e_x$ to get $n$ dollars from $e_y$, gets$(e_x, e_y, nNS)$ means 
for $e_x$ to get $n$ Nintendo Switch from $e_y$,
and negotiates$(e_x, e_y)?$ means
for $e_x$ to negotiate over the price of 1 Switch 
with $e_y$. 

Initially, either $e_2$, 
as in \textbf{Initial Step A} 
in \fbox{A}, 
or $e_3$, as in \textbf{Initial Step B},
gets 1 Switch from $e_1$ 
for \$300. 
One of the following holds in the end. 
\begin{enumerate}
    \item $e_2$ gets 2 Switch 
    from $e_1$ for \$600, and $e_3$ gets none. 
    \item $e_2$ gets 1 Switch 
    from $e_1$ for \$300, 
    $e_3$ then gets 1 Switch 
    from $e_1$ for \$300, and $e_3$ tries to sell 
    it to $e_2$ for \$400.  
    Since $e_2$ at that point has
    only \$350, he tries 
    to haggle over the price, to 
    settle at \$350. 
    \item $e_3$ gets 1 Switch 
    from $e_1$ for \$300, 
    $e_2$ then gets 1 Switch 
    from $e_1$ for \$300, and $e_3$ 
    tries to sell 
    it to $e_2$ for \$400.  
    Since $e_2$ at that point has
    only \$350, he tries 
    to haggle over the price, to 
    settle at \$350. 
    \item $e_3$ gets 2 Switch 
    from $e_1$, and tries to sell 
    them to $e_2$ for the total price of \$800. $e_2$ gets 1 Switch 
    from $e_3$ for \$400,
    but for the second Switch, 
     $e_2$ haggles
    over the price, to settle at \$250, 
    which fails due to $e_3$'s 
    preference. 
\end{enumerate}   
\noindent Whereas, if a bilateral negotiation
can progress uninterrupted till the end,  
the resource reallocation 
that the two agents negotiate over 
will be based on the resource allocation 
at the inception of the negotiation, 
a common business negotiation process is often not bound 
by the tight protocol guaranteeing no interruptions
\cite{Kim03,Sim13}. 
In the above example, 
even if $e_2$ wishes to buy 2 Switch consoles 
from $e_1$ as in the situation 1., $e_3$ may pick up  
a Switch from a shelf
by the time $e_2$ manages to take one of them, leading 
to the situation 2. or 3..   

The argumentation-based negotiation methodologies 
considered in the literature, mostly for two-party -
and not for interleaved - negotiations,
do not immediately 
scale up to addressing this concurrency issue. To adapt to it, in this work 
we propose a novel argumentation-based negotiation framework. We base the development 
on abstract persuasion argumentation (\APA) \cite{ArisakaSatoh18} which is 
a dynamic argumentation formalism that conservatively
extends Dung abstract argumentation \cite{Dung95}.
It accommodates a dynamic relation called persuasion 
    in addition 
to attack relation (an argument $a_1$ attacks 
an argument $a_2$) of abstract argumentation. Such a relation 
as: an argument $a_1$ 
converts an argument $a_2$ into another 
argument $a_3$ is expressible in \APA.
Also, it extends the notion of defence against an attack in Dung argumentation 
to a persuasion. Changes in resource allocation 
      as a result of a successful negotiation 
       can be modelled 
      as a successful persuasion. Failure of 
      a successful negotiation can be 
      modelled as an unsuccessful persuasion,  
      on the other hand.
      
We introduce the following new features 
to \APA.
\begin{itemize}
   \item \textbf{Quantification of arguments}:  
       We express a resource 
       as a quantified argument. 
       To describe
       that there are 2 Nintendo Switch,  
       we will have an argument: 
       there is a Nintendo Switch, 
       to which we assign 
       a numerical value 2 indicating the 
       quantity of the resource. 
  \item \textbf{Handshakes between persuasions}: 
    an agent often tries to obtain 
   a resource from another agent 
   who requires 
   a resource from the first agent in exchange. 
   While the two relations 
   may each be expressed as a persuasion in
   \APA, it has 
    no explicit mechanism to 
   enforce
   two persuasions to be considered 
   always together for 
   a dynamic transition. As a resolution, we introduce a handshake function. 
   \item \textbf{Numerical constraints for 
     attacks and persuasions}: 
     It matters in negotiation 
     how many (much) of one of the resources 
     are to be exchanged for how many (much)  
     of the other resource. It can also happen 
     that an agent attacks another agent's 
     argument with its argument 
     only under a certain condition. To intuitively
     express them, we allow both attacks and persuasions to be given numerical constraints.
   \item \textbf{Multi-agency}: each agent has its own argumentation 
   scope, and 
   may also apply its own criteria 
   to choose acceptable arguments within it. We 
   explicitly introduce 
   the notion of agents into \APA. 
\end{itemize}  
\subsection{Key contributions}  
\begin{itemize} 
\item \textbf{Argumentation-based 
  negotiation for interleaved negotiations}:  
   As far as we are aware, this is the first 
   negotiation theory in the line of abstract argumentation 
   \cite{Dung95} 
   that adapts to 
   interleaved multi-agent negotiations over 
   scarce resources.  
\item \textbf{Finer control over 
dynamic transitions}:  While {\APA} dynamics is already Turing-complete \cite{Arisaka19}, it does not 
mean that \APA's interpretation 
of transition relation \cite{ArisakaSatoh18} (detail 
is in section 2.2) can 
deterministically simulate a handshake of 
two persuasions for a transition. 
Numerical {\APA}
consolidates the synchronisation, 
accommodating a finer control 
over how a dynamic transition takes place. Also, it is rather cumbersome to have 
to deal 
with quantities without explicit use of numbers. 
Numerical {\APA} is an improvement on this aspect. 
\item \textbf{Explainable preference}: 
     In relevant theories (e.g. 
     \cite{Amgoud07,Hadidi10,Dung08,Kakas06}), 
     the use of preference relation 
     \cite{Bench-Capon03,Amgoud02} 
     over arguments to control 
     directionality of the attack (and other) relations 
      is ubiquitous.  
     However, the preference is often external to the underlying argumentation 
     with no concrete explanation given as to what formulates it. 
     In this work,  
     we instead rely upon numerical constraints to achieve 
     conditionalisation of attack (and persuasion) relations. Since
     they concern 
     quantities of resources, we can understand more explicitly 
     why some attack (persuasion) may not be present.    
\end{itemize} 
\subsection{Related work}  
Studies  
around concurrent negotiations are emerging, e.g. \cite{Niu18,Zhang17}. However, the problems we described above 
with argumentation-based approaches have not been 
considered in the literature, as far as our awareness goes.

Argumentation-based negotiation (Cf. \cite{Rahwan03} for an early survey) 
was proposed to obtain rational 
explanations as to why agents have or have not 
come to a deal. 

In \cite{Rahwan07}, a kind of agents' interaction during 
a bilateral negotiation through common goals was studied, 
under the assumption that there be no resource competition among agents.  
While we do not explicitly study cooperation among agents in this work, 
we allow resource competition, as was illustrated early in this section.  
We further allow interleaving of bilateral negotiations.

In \cite{amgoud11,Amgoud07}, agents  
negotiate over potential 
goals. Say, negotiation is over 
resource allocation, then 
each goal describes a possible resource 
allocation, represented as an argument. 
At each negotiation turn, new arguments 
and relations among them (such as attacks 
and supports) may be inserted 
into agents' scopes, to assist them 
make judgement as to which goal(s) should 
be accepted, with their preferences. There are other studies, 
e.g. \cite{Bonzon12}, that 
focus on some specific aspects of the theoretical framework. 
These works do not consider 
interleaved negotiations, however, for 
which the use of the goal arguments 
can be in fact problematic, as we are 
to illustrate later in section 3.1. Also, 
we approach towards agents' preferences 
via numerical information, as we described above. 

In \cite{Dung08}, a structural argumentation \cite{Dung09} 
was applied to contract-negotiations, where
features express properties of 
items including quantities. Yet again, 
it does not cover interleaved negotiations.

In the rest, we will present: technical preliminaries 
(in Section 2);  numerical 
abstract persuasion argumentation with the above-introduced
features 
(in Section 3); and examples (in Section 4),
before drawing conclusions.
\section{Technical Preliminaries} 
\subsection{Abstract Argumentation} 
Let $\mathcal{A}$ be a class of abstract entities that we understand 
as arguments, whose member is referred to by $a$ with or without a subscript, 
and whose finite subset is referred to by $A$ with or without a subscript. A finite argumentation framework is a tuple $(A, R)$  
with a binary relation over $A$ \cite{Dung95}. 
In this work, we assume $A \not= \emptyset$.
We denote the set of all finite argumentation 
frameworks by $\mathcal{F}$. 

The following definitions apply to any 
$(A, R) \in \mathcal{F}$. 
$a_1 \in A$ is said to attack $a_2 \in A$ if and only if, or iff, 
$(a_1, a_2) \in R$. 
 $(a_1, a_2) \in R$ 
is drawn graphically as $a_1 \rightarrow a_2$. 
$A_1 \subseteq A$ is said to defend $a_x \in A$ iff 
each $a_y \in A$ attacking  
$a_x$ is attacked by at least one member   
of $A_1$. $A_1 \subseteq A$ is said to be: 
    conflict-free iff   
        no member of $A_1$ attacks a member of $A_1$; 
    admissible iff 
        it is conflict-free and it defends 
        all the members of $A_1$; 
 complete iff it is admissible and  
        includes any argument it defends; 
     preferred iff   
        it is a set-theoretically maximal 
        admissible set; and 
   grounded iff it is the set intersection 
        of all complete sets of $A$.

Let $\Sem$ be  $\{\co, \pr, \gr\}$, 
and let $\Dung$ be $\Sem \times \mathcal{F} \rightarrow 2^{2^{\mathcal{A}}}$. 
Given a finite argumentation framework $(A, R)$, we denote by $\Dung(\semsmall, (A, R))$ the set of 
all $(A, R)$'s: complete sets iff $\semsmall = \textsf{co}$; 
preferred sets iff $\semsmall = \textsf{pr}$; 
and grounded sets iff $\semsmall = \gr$. 
By definition, $|\Dung(\gr, (A, R))| = 1$. There are many other 
types of semantics, but the three 
are quite typical semantics related 
(roughly) by complete sets. 

\subsection{$\APA$: Abstract Persuasion Argumentation}
Let $2^{(A, R)}$ denote 
$\{(A_1, R_1) \ | \ A_1 \subseteq A\ \andC\ R_1 = R \cap (A_1 \times A_1)\}$.\footnote{``$\andC$'' instead of 
 ``and'' is used when the context in which the word appears 
 strongly indicates truth-value comparisons. It follows the semantics 
 of classical logic conjunction.}  
$\APA$ \cite{ArisakaSatoh18} is a tuple  
$(A, R, \Rp, \Ainit, \hookrightarrow)$ 
for $\Ainit \subseteq A$; 
$\Rp \subseteq   
 A \times (A \cup \{\epsilon\}) \times 
A$; and 
 $\hookrightarrow: 2^A \times (2^{(A, R)} \times 2^{(A, R)})$. It extends 
 abstract argumentation $(A, R)$ conservatively. 
 $(a_1, \epsilon, a_2) \in \Rp$ 
is drawn graphically as $a_1 \multimap a_2$;
and $(a_1, a_3, a_2) \in \Rp$ as 
$a_1 \dashrightarrow a_3 \overset{a_1}{\multimap} a_2$.  \\
% $a_1 \in A$ is said to: attack $a_2 \in A$ iff $(a_1, a_2) \in R$, which 
% is drawn graphically as $a_1 \rightarrow a_2$; 
% induce $a_2 \in A$ iff $(a_1, \epsilon, a_2) \in \Rp$, 
% which is drawn graphically as $a_1 \multimap a_2$;
% and convert $a_2 \in A$ into $a_3 \in A$ iff 
% $(a_1, a_3, a_2) \in \Rp$, which is drawn graphically as 
% $a_1 \dashrightarrow a_3 \lol{\!a_1} a_2$.
\indent The following 
definitions apply to any $\APA$ tuple
$(A, R, \Rp, \Ainit, \hookrightarrow)$. 
Let $F(A_1)$ for some $A_1 \subseteq A$
denote $(A_1, R \cap (A_1 \times A_1))$.
% and let 
% $2^{(A, R)}$ denote 
% $\bigcup_{A_1 \subseteq A}F(A_1)$. For any $A_1 \subseteq A$, 
$F(A_1)$ for any $A_1 \subseteq A$ is said to 
be a state. $F(\Ainit)$ is called the initial state in particular. 
In any state $F(A_x)$, any member of $A_x$ is said to be visible 
in $F(A_x)$, while the other members of $A$ are said to be invisible in $F(A_x)$. 
A state $F(A_1)$ is said to be reachable iff $F(A_1) = F(\Ainit)$ or else there is some 
$F(A_2)$ such that $F(A_2)$ is reachable and that 
$(A_x, (F(A_2), F(A_1))) \in\ 
\hookrightarrow$, written
alternatively as $F(A_2) \hookrightarrow^{A_x} F(A_1)$ 
for some $A_x \subseteq A$. 
Such $A_x$ the transition is dependent upon 
was termed a reference set in \cite{ArisakaSatoh18},
but to make clearer its association 
to an agent, we call it an {\it agent set} instead. \\
\indent  $a_1 \in A$ is said to attack $a_2 \in A$ in a state $F(A_1)$ iff 
 $a_1, a_2 \in A_1$ $\andC$ $(a_1, a_2) \in R$.
 For $a_1, a_2, a_3 \in A$, 
 $a_1$ is said to be: inducing $a_3$ in a state $F(A_1)$ 
 iff $a_1 \in A_1$ $\andC$ $(a_1, \epsilon, a_3) \in \Rp$; and converting $a_2$ into $a_3$ in $F(A_1)$ iff 
 $a_1, a_2 \in A_1$ $\andC$ $(a_1, a_2, a_3) \in \Rp$.\\
 \indent $(a_1, \alpha, a_3) \in \Rp$ for $\alpha \in A \cup 
 \{\epsilon\}$ is said to be possible 
 in $F(A_1)$ with respect to an agent set 
 $A_x \subseteq A$ iff $a_1$ is either inducing 
 $a_3$ or converting $\alpha$ into $a_3$ in $F(A_1)$ 
 $\andC$ 
 $a_1$ is not attacked by any member of $A_x$ in $F(A_1)$.  
The set of all members of 
$\Rp$ that are possible in $F(A_1)$ with respect to an agent set 
$A_x \subseteq A$ is denoted by $\Gamma^{A_x}_{F(A_1)}$.
\subsubsection{Interpretation of $\hookrightarrow$.} The interpretation 
given of $\hookrightarrow$ in \cite{ArisakaSatoh18} is: (1) 
any subset $\Gamma$ of $\Gamma^{A_x}_{F(A_1)}$ can be simultaneously 
considered for transition from $F(A_1)$ into some $F(A_2)$; (2) 
if $\Gamma \subseteq \Gamma^{A_x}_{F(A_1)}$ is considered for transition 
into $F(A_2)$, then: (2a) if either $(a_1, \epsilon, a_3)$ or 
$(a_1, a_2, a_3)$ is in $\Gamma$, then 
$a_3 \in A_2$; (2b) if 
$(a_1, a_2, a_3)$ is in $\Gamma$, then 
$a_2$ is not in $A_2$ unless it is judged to be in $A_2$ by (2a); 
and (2c) if $a \in A_1$, then $a \in A_2$ unless 
it is judged not in $A_2$ by (2b). \\
\indent In other words, for $A_1 \subseteq A$  
and for $\Gamma \subseteq \Rp$,  
let $\nve^{A_1}(\Gamma)$ 
   be $\{a_x \in A_1 \ | \ \exists a_1 \in A_1\ 
   \exists a_3 \in A.(a_1, a_x, a_3) 
    \in \Gamma\}$, 
   and let $\pve^{A_1}(\Gamma)$  
   be $\{a_3 \in A \ | \ \exists a_1, \alpha \in A_1 \cup 
   \{\epsilon\}.\linebreak(a_1, \alpha, a_3)
   \in \Gamma\}$. 
   For $A_x \subseteq A$, $F(A_1)$ and $F(A_2)$, we have:
   $F(A_1) \hookrightarrow^{A_x} F(A_2)$ iff
   there is some $\emptyset \subset \Gamma \subseteq 
\Gamma^{A_x}_{F(A_1)} \subseteq \Rp$
   such that 
   $A_2 = (A_1 \backslash \nve^{A_1}(\Gamma))
    \cup \pve^{A_1}(\Gamma)$. 
\subsubsection{State-wise 
acceptability semantics.}  
$A_1 \subseteq A$ is said to be conflict-free 
in a (reachable) state $F(A_a)$ iff 
no member of $A_1$ attacks 
a member of $A_1$ in $F(A_a)$. 
%\textbf{Mutation-freeness}  We say that $A_1 \subseteq A$  
%is 
%mutation-free in a state $F(A_x)$ 
%iff for no reference set $A_u \subseteq A$ 
%there is a state $F(A_y)$ such that 
%$F(A_x) \hookrightarrow^{A_u} F(A_y)$ 
%and that $A_x \not= A_y$. \\
$A_1 \subseteq A$ is said to defend $a \in A$
in $F(A_a)$ from attacks iff, (if  $a \in A_a$, 
then) every $a_u \in A$ attacking $a$ 
in $F(A_a)$ is attacked 
by at least one member of $A_1$ in $F(A_a)$ (counter-attack).
 $A_1 \subseteq A$ is said to be proper 
in $F(A_a)$ 
iff $A_1 \subseteq A_a$. 
In \cite{ArisakaSatoh18}, there is another
criterion of no-elimination:  $A_1 \subseteq A$ is
said to defend $a \in A$ in $F(A_a)$ from 
eliminations iff, if $a \in A_a$, there is no state $F(A_b)$ 
such that both $F(A_a) \hookrightarrow^{A_1} F(A_b)$ 
and $a \not\in A_b$ at once (no elimination). 
Such condition can be made 
general: 
$A_1 \subseteq A$ is said to 
defend $a \in A$ in $F(A_a)$ 
from $k$-step eliminations, 
$k \in \mathbb{N}$, iff, if 
$a \in A_a$, then 
for every $0 \leq i \leq k$ 
and for every $F(A_a) \underbrace{\hookrightarrow^{A_1} 
 \cdots 
\hookrightarrow^{A_{1}}}_{i} F(A_{i})$, 
it holds that 
$a \in A_i$. \\
\indent $A_1 \subseteq A$ is said to be: 
$k$-step admissible 
in $F(A_a)$ iff $A_1$ is conflict-free, proper and
defends every member of $A_1$ in $F(A_a)$ from 
attacks and $k$-step eliminations; 
complete in $F(A_a)$ iff $A_1$ is admissible and 
includes every $a \in A$ it defends in $F(A_a)$ 
from attacks and $k$-step eliminations; preferred 
in $F(A_a)$ iff $A_1$ is maximally $k$-step complete 
in $F(A_a)$; 
and grounded in $F(A_a)$ iff it is the set intersection of all $k$-step complete sets 
in $F(A_a)$.   

In the rest of the paper, 
we assume $k$ to be always 0, for 
technical simplicity. 

\subsection{Abstract Argumentation with Agents}   
In the context of argumentation-based 
negotiations, it is common to consider
argumentation per agent. 
We can extend 
abstract argumentation $(A, R)$ with 
$A \not= \emptyset$ into 
$(A, R, E, \fe, \fsem)$ with: 
$E$ a set of agents;
a function $\fe: E \rightarrow (2^A \backslash 
\emptyset)$ which is such that, 
for $e_1, e_2 \in E$, 
if $e_1 \not= e_2$, 
then $\fe(e_1) \cap \fe(e_2) = \emptyset$; 
and $\fsem: E \rightarrow \Sem$. 
Intuitively, any $a \in \fe(e)$ is understood 
to be in the scope of the agent $e \in E$, 
and each agent $e \in E$ has its own 
semantics $\fsem(e)$. 
\section{Numerical Abstract Persuasion Argumentation}     
As per our discussion in Section 1, 
we extend $\APA$ with (1) numerical 
information, (2) a mechanism of 
handshakes, and (3) multi-agency. Prior 
to formally defining the theory, we 
provide intuition 
for the first two components through examples. 
\subsection{Numbers and numerical constraints}
Numbers and numerical constraints as always
are powerful enrichment 
to a formal system. In the context of this paper,
the following 
are the merits of having them. 
\subsubsection{Numbers help us better 
organise scarce resource allocations.}
Without 
them, it takes some effort to describe 
availability of resources in an argument. 
With goal arguments \cite{amgoud11}, 
each of all possible allocations of resources 
presently negotiated over 
is defined as an argument. For example, 
if $e_1$ and $e_2$ 
commence their negotiations over 
2 Switch and \$650, 
the allocation that 
2 Switch and \$0 are with $e_1$ and 0 Switch 
and \$650 
are with $e_2$ is one argument $a_u$, 
while the allocation that 
2 Switch and \$0 are with $e_2$ and 0 Switch 
and \$650 
are with $e_1$ is another argument $a_v$, 
and similarly for all the other possibilities. 
Bundling together every resource location 
in one big argument is not necessarily 
desirable in concurrent 
multi-agent negotiations, as availability of resources 
can change in the middle of the negotiation 
due to interruption by another negotiation, 
say between $e_1$ and $e_3$.
Such a change can entail disappearance 
of some of resource allocations that 
were initially feasible.

With numbers, the need for  
maintaining 
those big arguments is 
precluded. To describe 
the idea, 
we can have an argument {\it There 
is a Switch.} in the scope of $e_1$ (the 
argument $a_1$)
and an argument {\it There is a Switch.} 
in the scope of $e_2$ (the argument $a_2$).  
Similarly, we can have an argument 
{\it There is a dollar.} in the scope
of $e_1$ (the argument $a_3$) 
and {\it There is a dollar.} in the scope 
of $e_2$ (the argument $a_4$).  
Then, we can have a set $\Na$ of partial functions
$\mathcal{A} \rightarrow \mathbb{N}$ mapping 
an argument for a resource 
into its quantity in non-negative 
integer, 
e.g. $\na \in \Na$ can 
be such that $\na(a_1) = 2$ ($e_1$ has 2 Switch), 
$\na(a_2) = 0$ ($e_2$ has 0 Switch), $\na(a_3) = 0$ 
($e_1$ has \$0) and 
$\na(a_4) = 650$ ($e_2$ has \$650). 
These numerical information let 
us infer the earlier-mentioned big argument $a_u$. 
When there is any change to the quantities 
of the resources, we can replace $\na$ with 
another $\na_1 \in \Na$ which differs 
from $\na$ in mapping only 
for those resources (as arguments) affected 
by it. In the meantime, arguments themselves 
will stay the same. 
Making the overall resource allocation 
an inferrable - rather than hard-coded - information makes 
adaptation to interleaved negotiations  
simpler and more intuitive.   

\subsubsection{Numerical constraints help 
us express conditional attacks and persuasions.} 
Whether an agent attacks or 
attempts a persuasion (such 
as to solicit a concession of an offer) in a 
state could very well depend 
on the specific resource 
allocation in the state. An illustrative 
example is: $e_2$ thinks of 
\$300 already too expensive for an electronics
device, but, say the money in his possession 
is \$350, at least not so much as him being unable 
to buy a Switch. Thus, according 
to his preference (see on the second page of 
this paper), he is not happy 
but he does not complain (attack) at 
the seller for the ask price. This  
changes clearly when the ask price is 
\$400, since he would not be able to 
purchase it out of his pocket. There, the 
chance is that he actually complains 
(attacks) at the seller. 

Numerical constraints can help model 
this situation intuitively by allowing 
attacks to conditionally occur. 
Assume: 
\begin{itemize} 
\item Three arguments: {\it Dollars required.}  
in $e_1$'s scope 
($a_1$); 
{\it There is a dollar.} in $e_2$'s scope 
($a_2$); and {\it That costs a lot.} in     
$e_2$'s scope 
($a_3$).
\item $\na(a_1) = n_1$ and 
$\na(a_2) = n_2$.  
\item An attack $a_3\rightarrow a_1$ 
with a numerical constraint 
$n_2 < n_1$ given to it. 
\end{itemize}   
What these intend is that 
$a_3$ attacks $a_1$ when 
$e_1$'s ask price $n_1$ exceeds 
dollars $n_2$ in $e_2$'s possession. 
Numerical constraints can be similarly  
used for a persuasion, not only for 
an attack.

Whether they are 
for an attack or a persuasion, constraints 
may be left unspecified if irrelevant. However, 
it is rather inflexible to just allow 
constant numbers in a constraint, 
since we may like to express a condition: 
{\it if the quantity of $a_1$ is greater 
than that of $a_2$}, which depends on 
a chosen $\na \in \Na$. We therefore 
define a formal object for expressing 
a constraint, its synatx and semantics: 
\begin{definition}[Constraint objects: syntax]\label{def_numerical_objects_syntax} 
  Let $\nums, \expr$, and $\num$ with or without 
  a subscript be recognised 
  in the following grammar.  
  We assume $n$ to be a member of $\mathbb{N}$, 
  $a$ to be a member of $\mathcal{A}$, 
  and $r$ to be some $(a_1, a_2, a_3)$ with $a_1, a_2, a_3 \in 
  \mathcal{A}$. 
  Any $\nums$ (with or without 
  a subscript) recognised in this grammar is 
  assumed to be a finite set, and is called 
   a constraint object.  \\ 
  
  \noindent ${\ }\qquad\qquad\qquad\qquad \nums := \nums, \expr
  \ | \ \emptyset$.\\ 
  \noindent ${\ }\qquad\qquad\qquad\qquad \expr := \num = \num \ | \ \num < \num$.\\
  \noindent ${\ }\qquad\qquad\qquad\qquad \num := n \ | \ \star\! a \ | \ 
  \star\! r$.  \\

 %  \noindent Denote the set union of all $\nums$ recognised 
  % in this grammar by $2^{\nums}$, 
  % and let $\Pi$ with or without 
 %  a subscript refer to a finite subset of $2^{\nums}$. 
%   and the set of all $\num$ with an empty set 
%   $\emptyset$ added to it by $S^{\num}$,   
   \noindent 
   Let $S^{\nums}$ be the class of 
   all constraint objects, 
   and let $g: S^{\nums} \rightarrow 
    S^{\nums}$ be such that
   %Let $2^{\nums^G}$ be a subclass of $2^{\nums}$  
   %which is such that 
   $g(\nums) = \nums$ 
   iff no $\star a$
   for some $a \in \mathcal{A}$
   or $\star r$ for some 
   $r \in \mathcal{A} \times \mathcal{A} \times \mathcal{A}$
   occurs in $\nums$.  
   We denote the subclass of $S^{\nums}$ 
   that contains all $\nums \in S^{\nums}$ 
   with $g(\nums) = \nums$ but nothing else 
   by ${S^{\nums}}^G$. 
   We call $\nums \in {S^{\nums}}^G$ a ground 
   constraint object. 
\end{definition}  
For the semantics, we define an interpretation function 
from $S^{\nums}$ to ${S^{\nums}}^G$. 
\begin{definition}[Interpretation of constraint objects]\label{def_interpretation_numerical_objects} 
Let $S^{\num}$ denote the set of all distinct $\num$, 
%with an empty set $\emptyset$ added to it, 
and let $\nr: (\mathcal{A} \times {(\mathcal{A} \cup \{\epsilon\})} \times \mathcal{A}) \rightarrow S^{\num}$ 
be a partial function such that 
$\nr(r)$, if defined, 
is some $n \in \mathbb{N}$ or $\star a$ 
for some $a \in \mathcal{A}$. 
   Let $\mathcal{I}^t: S^\num \times \Na \rightarrow 
   S^\num$ be such that 
   $\mathcal{I}^t(x, \na)$ is: 
   $x$ if $x \in \mathbb{N}$; 
 %  $\emptyset$ if $x = \emptyset$; 
   $\na(a)$ if $x = \star a$; $\nr(r)$ 
   if $x = \star r$ $\andC$ $\nr(r) \in \mathbb{N}$; 
   $\mathcal{I}^t(\nr(r), \na)$ if $x = \star r$ 
   $\andC$ $\nr(r)$ is $\star a$ for some $a \in 
   \mathcal{A}$; and undefined, otherwise.
   
   Let $\mathcal{I}: S^{\nums} \times 
   \Na \rightarrow {S^{\nums}}^G$ be 
   such that  
   $\mathcal{I}(\nums, \na)$ is  
   $\nums_1$, where $\nums_1$ is 
   as the result of replacing 
   every occurrence of $x \in S^{\num}$ 
   in $\nums$ with $\mathcal{I}^t(x, \na)$. 
   $\mathcal{I}(\nums, \na)$ is defined 
   iff $\mathcal{I}^t(x, \na)$ is defined 
   for every occurrence of $x \in S^{\num}$ 
   in $\nums$. 
%   $\nums$ except that 
%   any occurrence of $x \in S^{\num}$ in 
%   $\nums$ i
%   $x \in \nums$ iff $\mathcal{I}^t(x, 
%           \na) 
%           \in \nums_1$.

   For any $\nums \in S^{\nums}$ and 
   for any $\na \in \Na$, we say that 
   $\mathcal{I}(\nums, \na)$ is interpretation 
   of 
   $\nums$ with respect to $\na$. 
\end{definition}   
\begin{definition}[Constraint objects: semantics]{\ }\\
   We define a predicate 
   $\sat: S^{\nums} \times \Na \rightarrow 
  \{\textsf{true}, \textsf{false}\}$ to be such 
   that 
   $\sat(\nums, \na)$ iff, 
   for every ${n_1 < n_2} \in 
   \mathcal{I}(\nums, \na)$, 
    $n_2$ is greater than 
    $n_1$, 
   $\andC$ 
   for every ${n_1 = n_2} \in 
   \mathcal{I}(\nums, \na)$,
    $n_1$ is equal to 
    $n_2$. 
   
%   We define $\quantity: \info \times 
%   \Na$ to be such that 
%   $\quantity((\nums, x), \na)$ is: $\mathcal{I}{^t}(x, \na)$ if 
%   $\mathcal{I}^t(x, \na) \in \mathbb{N}$; undefined, otherwise. 
\end{definition} 

\noindent The purpose of $\sat$ is to judge if 
the numerical constraints  
in $\nums \in S^{\nums}$ are satisfied 
for a given $\na \in N^a$. 
% In this particular work, we suppose 
% that any numerical information $c \in \info$
% given to an attack is such that $\quantity(c)$ 
% is undefined. 
\subsubsection{Numbers help us 
express the quantity needed 
of a resource for a dynamic 
transition.} 
Suppose $e_2$ has \$650, 
and that $e_1$ tries to obtain 
\$300 from $e_2$. After a successful 
transaction, 
$e_2$'s budget decreases, not 
to 0, however.
This scenario is not concisely   
expressed in \APA: 
if (1) $\{(a_1, a_2, a_3)\} = \Rp$, (2) both $a_1$ and 
$a_2$ are 
visible in a state (see Section 2), 
and (3) $(a_1, a_2, a_3)$ is possible in the state, 
then in the next state $a_2$ is invisible. However, with numbers, suppose: 
\begin{itemize}
    \item Three arguments: 
    {\it Dollars required.} in 
    $e_1$'s scope ($a_1$); 
    {\it There is a dollar.} in 
    $e_2$'s scope ($a_2$); 
    and {\it There is a dollar.} 
    in $e_1$'s scope ($a_3$). 
    \item $\na(a_2) = 650$ and $\na(a_3) = 
      0$. 
    \item 
     $(a_1, a_2, a_3) \in \Rp$, 
     with $\nr$
     assigning $300$ to $(a_1, a_2, a_3)$, i.e. 
     $\nr((a_1, a_2, a_3)) = 300$, 
     signifying how many dollars $e_1$ will require. 
\end{itemize}
Then the conversion $(a_1, a_2, a_3)$, provided 
it is possible in a given state, will 
update 
$\na$ to $\na_1 \in \Na$ such that $\na_1(a_2) = 350$; and  
$\na_1(a_3) = 300$, without mandatorily   
eliminating $a_2$.

\subsection{Handshakes among 
persuasion relations} 
Suppose: {\it Dollar required.} 
in the scope of $e_1$ (argument $a_1$);  
{\it There is a dollar.} 
in the scope of $e_2$ (argument $a_2$); 
 {\it There is a dollar.} 
in the scope of $e_1$ (argument $a_3$);
{\it Switch required.} 
in the scope of $e_2$ (argument $a_4$); 
{\it There is a Switch.} 
in the scope of $e_1$ (argument $a_5$); and 
{\it There is a Switch.} 
in the scope of $e_2$ (argument $a_6$).

Suppose the following two persuasions over them:  
one is $(a_1, a_2, a_3)$, another is $(a_4, a_5, a_6)$. 
To enforce a handshake between them, 
we introduce a function 
$\hs: \Rp \rightarrow 2^{\Rp}$ such that
(1) if $r_2 \in \hs(r_1)$, then 
$r_1 \in \hs(r_2)$, 
and that (2) if $\hs(r_1) \not= \emptyset$,  
then it is not considered for transition 
unless there is some $r_2 \in \hs(r_1)$ 
that is considered for transition simultaneously 
with it (see Section 2 
for what it means for a persuasion 
 to be considered 
for transition). The interpretation 
of $|\hs(r_2)| > 1$ is that $r_2$ may be 
considered 
together with a member (and not all 
the members) of $\hs(r_2)$. 
\subsection{Numerical Abstract Persuasion 
Argumentation}  
Denote the class of all partial 
     functions $(\mathcal{A} \times (\mathcal{A} \cup \{\epsilon\}) \times \mathcal{A}) \rightarrow S^{\num}$ by $N^r$, every member 
     of which is such that, 
     if defined, the output is some $n \in \mathbb{N}$
     or $\star a$ for some $a \in \mathcal{A}$. 
% With the above considerations, we obtain:
\begin{definition}[Numerical 
Abstract Persuasion Argumentation]  
We define a Numerical Abstract 
Persuasion Argumentation (Numerical 
\APA) to be a tuple 
$(A, R, \Rp, E, \fe,\linebreak \fsem, \Ainit, \hs, 
\Rightarrow, \nainit, \nr, \cst)$, with: 
(1) 
$\hs: \Rp \rightarrow 2^{\Rp}$;  
(2) 
$\nainit \in \Na$; (3) 
$\nr \in N^r$; (4) 
 $\cst: (R \cup \Rp) \rightarrow 
 S^{\nums}$; 
  and (5) $\Rightarrow: 2^A \times 
  ((2^{(A, R)} \times \Na) \times 
   (2^{(A, R)} \times \Na))$. \\
 \indent All the others, i.e. 
 $A$, $R$, $\Rp$, $E$, $\fe$, 
 $\fsem$ and $\Ainit$ are as defined in Section 2, 
 i.e. $A \subseteq_{\text{fin}} \mathcal{A}$, 
 $R \subseteq A \times A$, 
 $\Rp \subseteq A \times (A \cup \{\epsilon\}) 
 \times A$, 
 $E$ a set of agents, 
 $\fe: E \rightarrow (2^A \backslash 
 \emptyset)$ with 
 $\fe(e_1) \cap \fe(e_2) = \emptyset$ 
 if $e_1$ is not $e_2$, 
 $\fsem: E \rightarrow \Sem$, and 
 $\Ainit \subseteq A$. We assume 
 $A \not= \emptyset$. 
\end{definition}  
The following definitions apply to 
any Numerical {\APA} 
$(A, R, \Rp, E, \fe, \fsem, \Ainit, 
\hs, \Rightarrow, \nainit, \nr, \cst)$. 

\begin{definition}[States] 
    Let $F(A_1, \na_1)$ for 
    some $A_1 \subseteq A$ 
    and some $\na_1 \in \Na$  denote
    $((A_1, R \cap (A_1 \times A_1)), 
    \na_1)$. 
    We call any such $F(A_1, \na_1)$ 
    a (Numerical \APA) state.  
    We call $F(\Ainit, \nainit)$ 
    the initial state in particular. 
    
    In any state $F(A_x, \na_x)$, we say  
any member of $A_x$ visible in 
$F(A_x, \na_x)$, while the other 
members of $A$ invisible in $F(A_x, \na_x)$. 

We say that  $F(A_1, \na_1)$ is 
reachable iff $F(A_1, \na_1) = F(\Ainit, 
\nainit)$ or else there is some 
$F(A_2, \na_2)$ such that $F(A_2, \na_2)$ 
is reachable and that 
$F(A_2, \na_2) \Rightarrow^{A_y} F(A_1, \na_1)$ 
for some $A_y \subseteq A$.
\end{definition} 
Exactly how $\Rightarrow$ 
is interpreted is left unspecified 
at this point, which will 
be detailed later in section  \ref{subsection_interpretation}.
\subsection{Restrictions}\label{subsection_restrictions}
In this work, we will restrict 
our attention to 
a subset of
all Numerical {\APA} tuples. 
Specifically, 
\begin{enumerate}
   \item We divide $A$ into 
   arguments that denote 
   resources, and the other arguments 
   as ordinary arguments. We assume 
   that no resource arguments 
   become ordinary arguments, 
   or vice versa via state transitions.  
   \item We assume any resource argument 
   with quantity 0 is invisible. In particular,
   there is no $a \in \Ainit$ such that 
   $\nainit(a) = 0$. 
   \item We assume that $\nr \in N^r$ does not update 
   by dynamic transitions. This seems reasonable, 
   however, since $\Rp$ and 
   updates on $\na$ together easily simulate updates on $\nr$. 
   \item We assume that 
   $\nr$ is defined at most 
   for members of $\Rp$ that 
   are conversions.  We assume that
   $\nr((a_1, a_2, a_3))$  for $(a_1,
    a_2, a_3) \in \Rp$ with $a_1, a_2, a_3 \in A$ 
    is defined iff 
     $a_2$ and $a_3$ are a resource argument. 
     This is for our intended purpose, 
     that
     $\nr((a_1, a_2, a_3))$ 
      represents 
     the number (amount) of 
     resources 
     $a_1$ asks of from $a_2$ which 
     changes the number (amount) of resources 
     $a_3$. We further assume 
     that if $\nr((a_1, a_2, a_3))$ for $(a_1, a_2, a_3) \in \Rp$ 
     is defined, then for every $a_4, a_5 \in A$, if 
     $(a_4, a_2, a_5) \in \Rp$, then 
     $\nr((a_4, a_2, a_5))$ is defined and vice versa. 
\end{enumerate}

\noindent Formal definitions follow.  
\begin{definition}[Resource and 
ordinary 
arguments]\label{def_resource_and_ordinary_arguments}  
    We say that 
    $a \in A$ is a resource argument in 
    $F(A_1, \na_1)$
    iff $\na_1(a)$ is defined.  
    We say that 
    $a \in A$ is an ordinary argument  
    in $F(A_1, \na_1)$
    iff $a$ is not a resource argument 
    in $F(A_1, \na_1)$. 
\end{definition} 
\begin{definition}[Type rigidity]\label{def_argument_type_regidity} 
    We say that $A$ is type rigid
    iff for every 
     $F(A_x, \na_x)$, 
    $a \in A$ is a resource argument 
    in $F(\Ainit, \nainit)$ iff 
    $a$ is a resource argument 
    in $F(A_x, \na_x)$. 
\end{definition} 
\begin{definition}[Normal relations]\label{def_standard_relation} 
    We say that $\Rp$ is normal 
    iff for every 
    $F(A_1, \na_1)$:
    \begin{itemize} 
    \item for every $(a_1, a_2, a_3) \in \Rp$ with 
    $a_1, a_2, a_3 \in A$, 
    $\mathcal{I}^t(\nr((a_1, a_2, a_3)), 
    \na_1)$ 
    is defined 
    iff $a_2$ and $a_3$ 
    are a resource argument in $F(A_1, \na_1)$.  
    \item If $\mathcal{I}^t(\nr((a_1, a_2, a_3)), 
    \na_1)$ for $(a_1, a_2, a_3) \in \Rp$ is defined, 
    then for any 
     $a_4, a_5 \in A$, $(a_4, a_2, a_5) \in \Rp$ iff 
    $\mathcal{I}^t(\nr((a_4, a_2, a_5)), \na_1)$ is defined. 
    \end{itemize} 
\end{definition}  

\noindent In the rest, we assume a restricted 
class of 
Numerical $\APA$ tuples with  
type-rigid $A$ and 
normal $\Rp$, and which are moreover 
such that 
$\nr$ remains constant through state transitions. 
\subsection{Attacks and persuasions} 
Some attacks and persuasions may not satisfy 
constraint objects attached to them 
in a state, 
in which case we simply ignore 
their influence in the state, 
as embodied in: 
\begin{definition}[Constraint-adjusted 
relations]
We say:
\begin{itemize}
\item  $R'$ 
is an attack relation constraint-adjusted 
in $F(A_1, \na_1)$ iff 
$R' = \{(a_1, a_2) \in {R \cap (A_1 \times 
A_1)} \ | \ 
\sat(\cst((a_1, a_2)), \na_1)\}$.  \\

\item $\Rp'$ is a persuasion 
relation constraint-adjusted in
$F(A_1, \na_1)$ 
iff all the three conditions 1., 2. and 3. hold good for every $r \in \Rp'$.
\begin{enumerate} 
   \item If $r \equiv (a_1, \epsilon, 
   a_3)$ for some $a_1, a_3 \in A$, then 
   $a_1 \in A_1$ $\andC$ 
   $\sat(\cst(r), \na_1)$. 
   \item If $r \equiv (a_1, a_2, a_3)$ 
   for some $a_1, a_2, a_3 \in A$, 
   then $a_1, a_2 \in A_1$ $\andC$ 
   $\sat(\cst(r), \na_1)$.  
   \item There
exists no $\Rp''$ such that 
$\Rp' \subset \Rp''$ and that 
$\Rp''$ satisfies both of the conditions 
1. and 2. above. 
\end{enumerate} 
\end{itemize}
We denote the attack relation 
constraint-adjusted in $F(A_1, 
\na_1)$ by  
${\inducedR(F(A_1, \na_1))} (\subseteq R \cap 
(A_1 \times A_1))$, and the 
persuasion relation constraint-adjusted
in $F(A_1, \na_1)$ by 
${\inducedRp(F(A_1, \na_1))} (\subseteq \Rp 
\cap (A_1 \times (A_1 \cup \{\epsilon\}) \times A))$. 
\end{definition} 
We use the constraint-adjusted 
relations to characterise 
attacks and persuasions 
in a state. 
\begin{definition}[Attacks and persuasions 
in a state]\label{def_attacks_persuasions}  
We say $a_1 \in A$ attacks $a_2 \in A$ 
in a state $F(A_1, \na_1)$  iff 
$(a_1, a_2) \in \inducedR(F(A_1, \na_1))$. 
We say $a_1 \in A$ is inducing $a_2 \in A$ 
in a state $F(A_1, \na_1)$
iff $(a_1, \epsilon, a_2) \in \inducedRp(F(A_1, \na_1))$.
We say $a_1 \in A$ is converting $a_2 \in A$ 
to $a_3 \in A$ in a state $F(A_1, \na_1)$  
iff 
$(a_1, a_2, a_3) \in \inducedRp(F(A_1, \na_1))$. 
\end{definition} 

\subsection{State-wise agent semantics} 
We say $A_1 \subseteq A$ is conflict-free
in 
$F(A_a, \na_a)$ iff 
no member of $A_1$ attacks 
a member of $A_1$ in $F(A_a, \na_a)$.
We say 
$A_1 \subseteq A$ defends $a \in A$ 
in $F(A_a, \na_a)$ from attacks 
iff, if $a \in A_a$, then 
every $a_u \in A_a$ attacking 
$a$ in $F(A_a, \na_a)$ 
is attacked by
at least one member of $A_1$ in 
$F(A_a, \na_a)$. 
We say $A_1 \subseteq A$  
is proper in $F(A_a, \na_a)$ 
iff $A_1 \subseteq A_a$. 

As was stated in Section 2, 
in this work we will not deal with 
defence from eliminations. 
With this simplification,
each agent semantics 
in a given state is 
knowable without consideration 
over dynamic transitions.    

\begin{definition}[State-wise 
agent admissibility]\label{def_state_wise_agent_semantics}
We say that
$A_1 \subseteq A$ 
is: admissible in $F(A_a, \na_a)$  
for $e \in E$ iff  
$A_1 \subseteq \fe(e)$ $\andC$ 
$A_1$ is conflict-free $\andC$  
$A_1$ is proper $\andC$ 
$A_1$ defends every member of 
$A_1$ from attacks in $F(A_a, \na_a)$; 
complete in $F(A_a, \na_a)$ 
for $e \in E$ iff
$A_1$ is admissible in $F(A_1, \na_1)$ for $e$
$\andC$ $A_1$ includes 
every $a \in \fe(e)$ it defends from attacks
in $F(A_a, \na_a)$; 
preferred in $F(A_a, \na_a)$ 
for $e \in E$ iff  
$A_1$ is a maximal complete 
set in $F(A_a, \na_a)$ for $e$; 
and grounded in $F(A_a, \na_a)$ 
for $e \in E$ iff 
$A_1$ is the set intersection 
of all complete sets in $F(A_a, \na_a)$ for $e$.  
\end{definition} 
Together with the choice of a semantic type $\fsem(e)$ by each agent $e \in E$, 
we obtain: 
\begin{definition}[State-wise agent semantics] {\ }\\
    Let $\AS: 2^A \times \Na \times E \rightarrow 
       2^{2^A}$ be such that 
    $\AS(A_1, \na_1, e_1)$ is 
      the set of all:
    \begin{itemize} 
       \item complete
         sets in $F(A_1, \na_1)$ for $e_1$ 
         if $\fsem(e_1) = \textsf{co}$. 
       \item preferred sets in $F(A_1, \na_1)$ for $e_1$ 
         if $\fsem(e_1) = \textsf{pr}$. 
       \item grounded sets in $F(A_1, \na_1)$ for $e_1$ 
         if $\fsem(e_1) = \textsf{gr}$. 
    \end{itemize}  
   For every $F(A_1, \na_1)$ and every $e \in E$, 
   we call $\AS(A_1, \na_1, e)$  
   $e$'s semantics in $F(A_1, \na_1)$. 
\end{definition} 

\subsection{Interpretation of $\Rightarrow$} \label{subsection_interpretation}
As in {\APA} (see Section 2), 
we then characterise 
 possible persuasions 
with respect to some agent set.     
\begin{definition}[Possible persuasions]\label{def_possible_persuasions} 
    $(a_1, \alpha, a_2) \in \Rp$ 
    for $\alpha \in A \cup \{\epsilon\}$ 
    is said to be possible 
    in $F(A_1, \na_1)$ with respect to 
    an agent set $A_x \subseteq A$ iff 
    there is some $e_1 \in E$ such that 
    $A_x \in \AS(A_1, \na_1, e_1)$ (note
    that $A_x$ may be an empty set; see Section 2) 
    $\andC$ 
    $a_1$ is either inducing $a_2$ or converting 
    $\alpha$ into $a_2$ in $F(A_1, \na_1)$ 
    $\andC$ $a_1$ is not attacked 
    by any member of $A_x$ in $F(A_1, \na_1)$.   
      We denote by $\Gamma^{A_x}_{F(A_1, \na_1)}$ 
    the set of all members of $\Rp$  
    that are possible in $F(A_1, \na_1)$ 
    with respect to an agent set $A_x \subseteq A$. 
\end{definition}  
There may be more than one agent in a Numerical {\APA}; 
some persuasions must always be together due to $\hs$; 
and, on the other hand, some persuasions cannot be together 
due to the quantity change of a resource otherwise 
going below 0.  We 
thus obtain a multi-agent version of Definition \ref{def_possible_persuasions}: 
\begin{definition}[Multi-agent possible persuasions]\label{def_multi_agent_possible_persuasions} 
    We say that $A_x \subseteq A$ is a multi-agent union set in $F(A_1, \na_1)$
    iff $(A_x \cap \fe(e))  \in \AS(A_1, \na_1, e)$ for every $e \in E$.   \\
    \begin{adjustwidth}{0.2cm}{} 
    (Explanation: every agent may have more than one set of arguments in 
    its semantics in $F(A_1, \na_1)$. While any one of them may be chosen by the agent, 
    not two distinct ones can they choose at the same time. Therefore, 
    when we obtain the set union of one member of 
    every agent's semantics, it should trivially hold that 
    the set intersection of the union set and $\fe(e)$ 
    is a member of $e$'s semantics in $F(A_1, \na_1)$.) \\
    \end{adjustwidth}  
    For each multi-agent union 
    set $A_x$ in $F(A_1, \na_1)$, 
    let $\Lambda_{F(A_1, \na_1)}^{A_x}$
    denote 
    $\bigcap_{e \in E} \Gamma^{A_x \cap \fe(e)}_{F(A_1, \na_1)}$ (denoting possible 
    persuasions with respect 
    to each $A_x \cap \fe(e)$ 
    ($e \in E$)), 
    we say that $(\emptyset \subset)\ \Lambda \subseteq 
    \Lambda_{F(A_1, \na_1)}^{A_x}$ is a multi-agent possible 
    persuasion set in 
    $F(A_1, \na_1)$ with respect to $A_x$ iff both of 
    the following conditions hold: 
    \begin{enumerate}
       \item For every  $r_1 \in \Lambda$, 
    if $\hs(r_1) \not= \emptyset$, 
    then there exists one and only one $r_2 \in \hs(r_1)$
    such that $r_2 \in \Lambda$. (\textbf{Handshake-compatibility}) 
       \item Let $\decreaser, \increaser: 2^A \times \Na \times 2^{\Rp} \times A \rightarrow \mathbb{N}$  
       be such that: 
     \begin{itemize} 
     \item 
       $\decreaser(A_1, \na_1, \Gamma, a) = \{(a_1, a, a_3) \in \Gamma \ | \  
      \nr((a_1, a, a_3)) \text{ is defined.}\}$. 
       \item $\increaser(A_1, \na_1, \Gamma, a) = \{(a_1, a_2, a) \in \Gamma \ | \ 
        \nr((a_1, a_2, a)) \text{ is defined.}\}$. 
      \end{itemize}    
   \begin{adjustwidth}{0.2cm}{} 
  (Explanation: 
       $\decreaser(A_1, \na_1, \Gamma, a)$ 
       returns the set of 
       conversions in $\Gamma$ on $a$, 
       and $\increaser(A_1, \na_1, \Gamma, a)$
       returns the set of 
       conversions 
       in $\Gamma$ that induce $a$.)\\
       \end{adjustwidth} 
       Then, for $a \in A_1$, 
       if $a$ is a resource argument, 
       $\na_1(a) + \Sigma_{r \in \increaser(A_1, 
       \na_1, \Lambda, a)}\ \mathcal{I}^t(\nr(r), \na_1)\\
    - \Sigma_{r \in \decreaser(A_1, \na_1, \Lambda, a)}\ \mathcal{I}^t(\nr(r), \na_1) \geq 0$. 
       (\textbf{Resource-safety}) 
    \end{enumerate} 
    % We denote by  $\Xi_{F(A_1, \na_1)}$ the least set that includes 
    % every multi-agent possible persuasion set in $F(A_1, \na_1)$ with respect to 
    % some multi-agent union set in $F(A_1, \na_1)$.
\end{definition}  
\noindent A multi-agent union set defined 
above involves every $e \in E$, 
to express defence by 
every agent against dynamic transitions with 
the set of arguments it chooses. 
\noindent Finally: 
\begin{definition}[Interpretation of $\Rightarrow$] 
We define: $F(A_1, \na_1) \Rightarrow^{A_x} F(A_2, \na_2)$ iff 
(1) $A_x$ is a multi-agent union set $\andC$ (2) 
there is some multi-agent possible persuasion set
$\Lambda$ in $F(A_1, \na_1)$ with respect to 
$A_x$ such that: 
\begin{itemize}
   \item 
      $\na_2(a) = \na_1(a) + \Sigma_{r \in \increaser(A_1, \na_1, \Lambda, a)}\ \mathcal{I}^t(\nr(r), \na_1) 
      - \Sigma_{r \in \decreaser(A_1, \na_1, \Lambda, a)}\ \mathcal{I}^t(\nr(r), \na_1)$ for every resource argument $a \in A$. 
  \item Let $\nve(\Lambda)$ and $\pve(\Lambda)$ be:
   \begin{itemize}
      \item $\nve(\Lambda) = \{{a_x \in A_1} \ | \ {\exists a_1 \in A_1}\ 
   {\exists a_3 \in A.}(a_1, a_x, a_3) 
    \in \Lambda \ \andC\\  
    {\ }\qquad\qquad\qquad\qquad\quad({\na_2(a_x) = 0} \ \orC\ \na_2(a_x) \text{ is undefined.})
    \}$. 
     \item $\pve(\Lambda) = \{a_3 \in A \ | \ \exists a_1, \alpha \in A_1 \cup 
   \{\epsilon\}.(a_1, \alpha, a_3)
   \in \Lambda \ \andC\\
   {\ }\qquad\qquad\qquad\qquad\quad ({\na_2(a_3) \not = 0} \ \orC\ \na_2(a_3) \text{ is undefined.})\}$.
   \end{itemize} 
     Then $A_2 = (A_1 \backslash \nve(\Lambda)) \cup \pve(\Lambda)$. 
\end{itemize}

\end{definition}

\section{Example Modelling of Concurrent Multi-Agent 
Negotiations}   
We illustrate Numerical {\APA} with 
our running example. 
We assume the following 
graphical conventions in all figures. 
\begin{itemize}
    \item $a \in A$ is generally 
    drawn as $a$. As an exception, 
    for 
    $a \in A$ with a 
    defined $\na_1(a)$ in $F(A_1, \na_1)$, 
    it may be drawn more 
    specifically as 
    $\na_1(a):a$, or as in any 
    form that puts 
    the quantity before the argument.  
    \item A visible argument is bordered, 
     an argument that is not visible is not 
     bordered. 
    \item $(a_1, a_2) \in R$  
    with $\cst((a_1, a_2)) = \nums$ is drawn 
    as 
    $a_1 \xrightarrow[]{\nums} a_2$. 
    $\nums$ may be omitted if $\nums = \emptyset$, 
    and brackets to indicate a set of 
    constraints 
    may be omitted if it is a singleton 
    set. 
    \item $(a_1, \epsilon, a_3) \in \Rp$ 
    with $\cst((a_1, \epsilon, a_3)) = 
    \nums$ is drawn as 
    $a_1 \overset{\nums}{\multimap} a_3$. $\nums$ 
    may be omitted if $\nums = \emptyset$, 
     and brackets may be omitted if it is a singleton 
    set.  
    \item $(a_1, a_2, a_3) \in \Rp$ 
    with $\cst((a_1, a_2, a_3)) = \nums$ 
    and a defined $\nr((a_1, a_2, a_3)) = x\ (\in S^{\num})$
     is drawn as 
    $a_1 \overset{\nums}{\underset{x}{\dashrightarrow}} a_2 \overset{a_1}{\multimap} 
    a_3$.  $\nums$ may be omitted if 
    $\nums = \emptyset$, 
     and brackets may be omitted if it is a singleton 
    set. If $\nr((a_1, a_2, a_3))$ is not defined, 
    then $x$ is not stated. 
\end{itemize}     
Let us reflect back on the negotiation example of 
Section 1, more specifically on the 
stage of the negotiation  
in \fbox{A} right after 
\textbf{Initial Step B}  
($e_1$ gives 1 Switch 
to $e_3$ for \$300)
is taken. 
There are 16 arguments in total that appear in 
at least one of the 3 figures below. 
Out of them, $a_{12-16}$ are 
$e_1$'s arguments, 
$a_{7-11}$ are $e_2$'s arguments, 
and $a_{1-6}$ are $e_3$'s arguments. 
\fbox{B}, \fbox{C}, and \fbox{D}
represent 
an argumentation-based 
bilateral negotiation for $e_3$(left
column)-$e_2$(right column), 
$e_1$(left column)-$e_2$(right column), and $e_1$(left column)-$e_3$(right column), respectively. 
%These figures are for the same state, 
%and those arguments with a border around them  
%are visible, while 
%all the others are invisible in the state.
Some attacks and persuasions may not be drawn 
in those figures if they are 
not contained within 
the bilateral negotiation. 
Formally, they represent 
the following Numerical \APA. 
What exactly $\nums_1, \nums_2$ and $\nums_4$ in \fbox{B} and 
\fbox{C} are are also stated below. 
\begin{figure}[!h] 
    \includegraphics[scale=0.16]{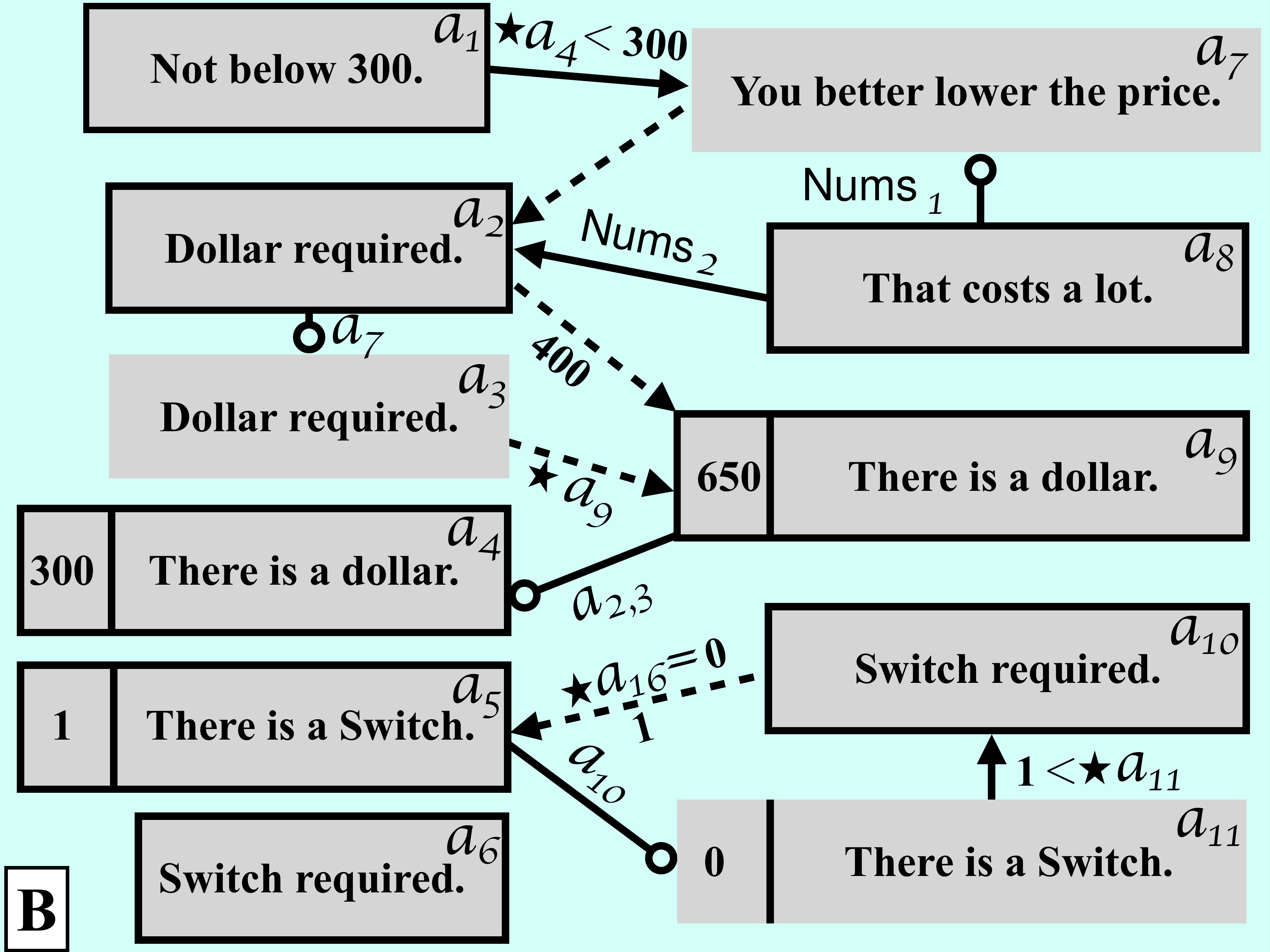} %e2-e3
    \includegraphics[scale=0.16]{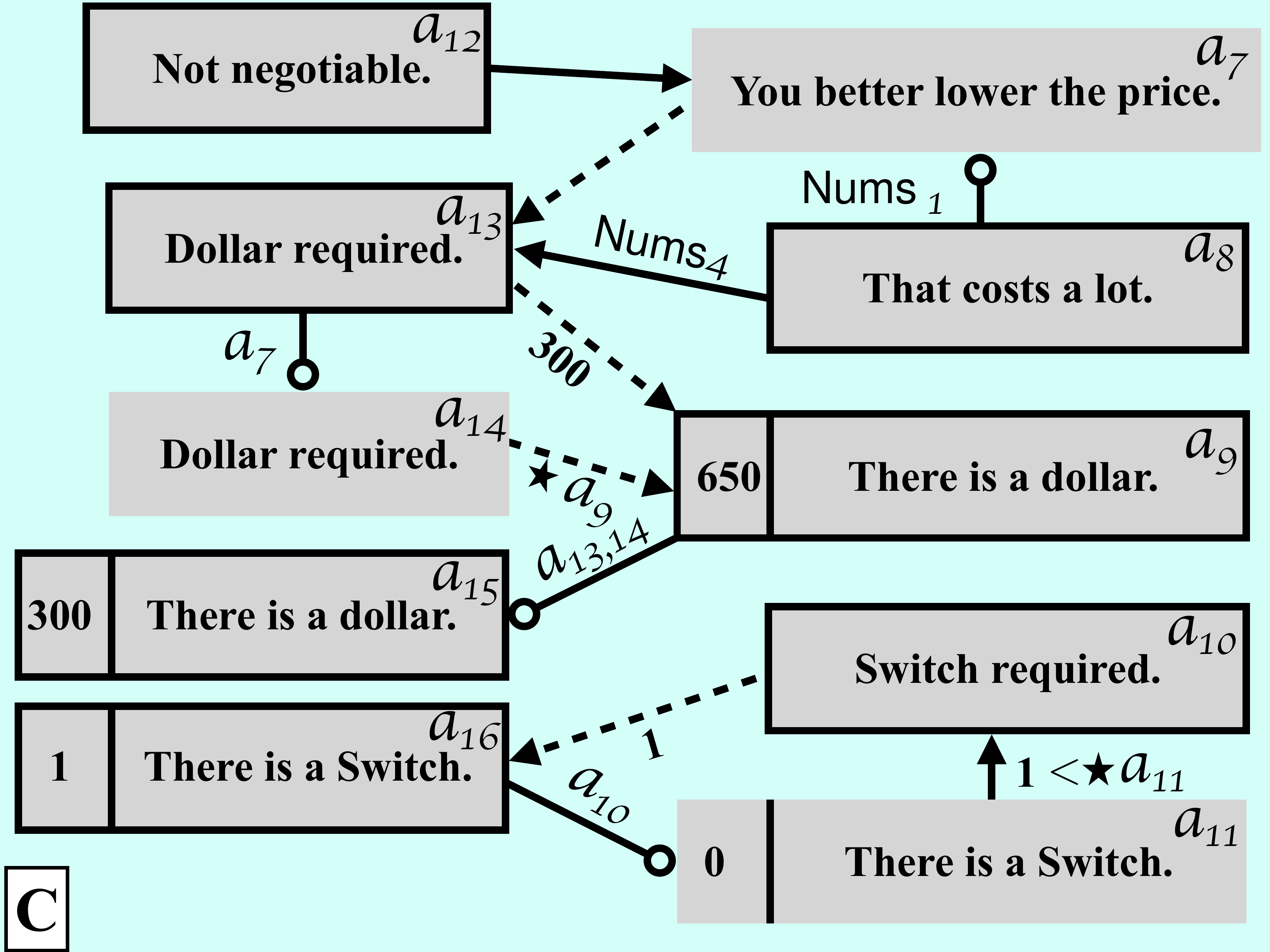} %e1-e2
    \includegraphics[scale=0.16]{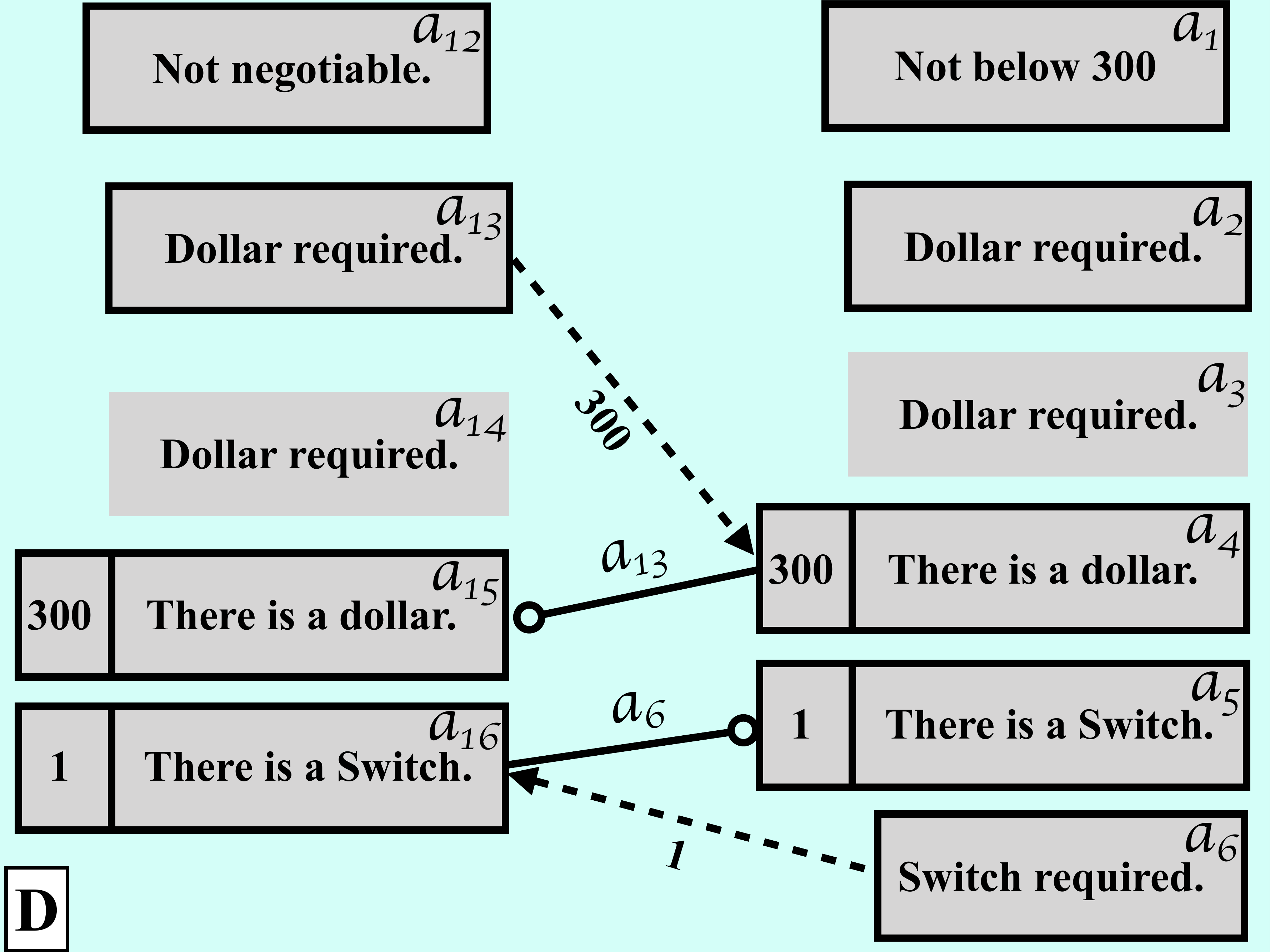} %e1-e3
    \includegraphics[scale=0.16]{blank.pdf} 
\end{figure}
{\small 
\begin{itemize}
   \item $A \equiv 
   \{a_1, \ldots, a_{16}\}$. 
   \item $E = \{e_1, e_2, e_3\}$.
   \item $R \ \equiv  \; 
     \{(a_8, a_2), \quad\qquad (a_1, a_7), \quad\qquad\ (a_{12}, a_7), \quad\qquad 
        (a_8, a_{13}), \quad\qquad (a_{11}, a_{10})\}$.
    \item $\Rp \equiv 
    \{
      (a_2, a_9, a_4), \ \ \quad (a_3, a_9, a_4), \qquad 
      (a_7, a_2, a_3),\\ 
       {\ }\qquad\quad (a_8, \epsilon, a_7), \qquad 
      (a_{10}, a_5, a_{11}),\ \quad 
      (a_7, a_{13}, a_{14}),\;   \quad 
      (a_{13}, a_{9}, a_{15}),\\ 
      {\ }\qquad\quad (a_{14}, a_{9}, a_{15}),\ \ \; 
      (a_{10}, a_{16}, a_{11}), \ \ \;
      (a_{13}, a_4, a_{15}), \quad (a_6, a_{16}, a_5)\}$. 
     
    \item $\fe(e_1) = \{a_{12}, \ldots, a_{16}\}$, \qquad 
    $\fe(e_2) = \{a_7, \ldots, a_{11}\}$, \qquad 
    $\fe(e_3) = \{a_1, \ldots, a_6\}$. 
    \item $\fsem(e_1) = \fsem(e_2) = \fsem(e_3) = \pr$ 
    (for this example, it does not matter 
    which of the three semantics each one is). 
    \item $\Ainit = \{a_1, \quad a_2, \quad a_4, \quad a_5, \quad a_8, \quad a_9, \quad   
    a_{10}, \quad a_{12}, \quad a_{13}, \quad  a_{15}, \quad  a_{16}\}$. 
    \item $\hs((a_2, a_9, a_4)) = 
         \{(a_{10}, a_5, a_{11})\}$, \qquad\qquad\quad \
         $\hs((a_3, a_9, a_4)) = 
         \{(a_{10}, a_5, a_{11})\}$, \\
         ${\hs((a_{10}, a_5, a_{11})) = 
         \{(a_2, a_9, a_4), (a_3, a_9, a_4)\}}$, 
         $\hs((a_{13}, a_9, a_{15})) =  
         \{(a_{10}, a_{16}, a_{11})\}$, 
        $\hs${\scriptsize$((a_{10}, a_{16}, a_{11})) = 
         \{(a_{13}, a_9, a_{15}), (a_{14}, a_9, a_{15})\}
         $},\ \  
         $\hs((a_{14}, a_9, a_{15})) = 
         \{(a_{10}, a_{16}, a_{11})\}$, 
         \\
         $\hs((a_{13}, a_{4}, a_{15})) = 
         \{(a_6, a_{16}, a_5)\}$, \qquad\qquad\quad 
         $\hs((a_6, a_{16}, a_5)) = 
         \{(a_{13}, a_4,a_{15})\}$.  
    \item $\Rightarrow$ is 
    as per section \ref{subsection_interpretation}.
    \item $\nainit(a_4) = 300, \nainit(a_5) = 1$,
    $\nainit(a_9) = 650$,
    $\nainit(a_{11}) = 0$, 
    $\nainit(a_{15}) = 300, \nainit(a_{16}) = 1$.  
    $\nainit$ is not defined for the other arguments.  
    \item {\small $\nr((a_2, a_9, a_4)) = 400$, \quad \quad 
    $\nr((a_3, a_9, a_4)) = \star a_9$}, \quad \quad 
    $\nr((a_{10}, a_5, a_{11})) = 1$, \\
    $\nr((a_{13}, a_9, a_{15})) = 300$,\ \  \quad 
    $\nr((a_{14}, a_9, a_{15})) = \star a_9$,\ \ \quad 
    $\nr((a_{10}, a_{16}, a_{11})) = 1$,\\ 
    $\nr((a_{13}, a_4, a_{15})) = 300$, \ \quad 
    $\nr((a_{6}, a_{16}, a_{5})) = 1$.   \\
    $\nr$ is not defined for any other members of $\Rp$. 
    \item $\cst((a_1, a_7)) =    
    \{\star a_9 < 300\}$.\\
        $\cst((a_8, a_2, a_3)) = \nums_2 =  
        \{\star a_9 < \star (a_2, a_9, a_4), \star (a_2, a_9, a_4) < 
        \star (a_{13}, a_9, a_{15})\}$.  \\
         $\cst((a_{8}, a_{13}, a_{14})) = \nums_4 =  
         {\{\star a_9 < 
   \star (a_{13}, a_9, a_{15}), \star (a_{13}, a_9, a_{15}) < 
   \star (a_2, a_9, a_4)\}}$. \\  
   $\cst((a_{11}, a_{10})) = \{1 < \star a_{11}\}$. \\
   $\cst((a_8, \epsilon, a_7)) = \nums_1 = \{{\star a_9 < 
   \star (a_{13}, a_9, a_{15}),}
        {\star a_9 < \star (a_2, a_9, a_4)}\}$. 
\end{itemize}   
}

\subsubsection{Preference as quantities 
and numerical constraints.}  
Both $e_1$ and $e_3$ are trying 
to sell 1 Nintendo Switch console that they each have 
to $e_2$, 
see $(a_{13}, a_9, a_{15}) \in \Rp$ 
in \fbox{C} for $e_1$-$e_2$ negotiation
and $(a_2, a_9, a_4) \in \Rp$ in \fbox{B} 
for $e_3$-$e_2$ negotiation. 
According to 
the preference spelled out 
in Section 1 (on the second page), $e_2$ chooses 
the cheapest offer. The 
preference is enforced in  
the constraint on the attack $a_8 \rightarrow 
a_{13}$, i.e. $\nums_4$ (\fbox{C}),
and that on the attack 
$a_8 \rightarrow a_2$, i.e. 
$\nums_2$ (\fbox{B}). Specifically,  
$\star (a_2, a_9, a_4) < \star (a_{13}, a_9, a_{15})$ in $\nums_2$ 
dictates that $e_2$ complains at 
$e_1$ of her ask price and does not 
consider the deal 
if $e_3$'s ask price is lower. 
Similarly, $\star (a_{13}, a_9, a_{15}) < \star (a_2, a_9, a_4)$ in $\nums_4$ 
dictates that $e_2$ complains at 
$e_3$ of his ask price and does not 
consider the deal 
if $e_1$'s ask price is lower.  
\subsubsection{Non-deterministic 
dynamic transitions for  
concurrent multi-agent 
negotiations.} 
As in the steps indicated in \fbox{A}, 
in this state, 
either $e_2$ or $e_3$, but not 
both of them, may obtain a Switch 
from $e_1$ (see \fbox{C} for $e_1$-$e_2$
negotiation, and \fbox{D} for $e_1$-$e_3$ 
negotiation). 
The reason 
that they cannot both obtain   
a Switch from $e_1$ is due to 
(\textbf{Resource-safety}) of Definition 
\ref{def_multi_agent_possible_persuasions}, 
since 
$\mathcal{I}^t((a_{10}, a_{16}, a_{11}), \nainit) 
+ \mathcal{I}^t((a_6, a_{16}, a_5), \nainit) = 2$
which is strictly greater than  
$\nainit(a_{16}) = 1$. However, 
otherwise, either 
$(a_{10}, a_{16}, a_{11})$ and 
$(a_{13}, a_9, a_{15})$ 
or else $(a_{6}, a_{16}, a_5)$ 
and $(a_{13}, a_4, a_{15})$ 
 may 
be considered together for 
a dynamic transition in $F(\Ainit, \nainit)$. 
To wit,
observe: 
{\small 
\begin{itemize}
   \item $\AS(\Ainit, \nainit, e_1)
   = \{\{a_{12}, a_{13}, a_{15}, a_{16}\}\} \equiv \{A_x\}$. \\
   $\AS(\Ainit, \nainit, e_2) = \{\{a_8, 
   a_9, a_{10}\}\} \equiv 
   \{A_y\}$. \\
   $\AS(\Ainit, \nainit, e_3) = \{\{a_1, a_2, 
   a_4, a_5, a_6\}\} \equiv \{A_z\}$. 
   \item $\Gamma^{A_{i}}_{F(\Ainit, \nainit)} 
   = \{(a_{13}, a_4, a_{15}), 
      (a_{6}, a_{16}, a_5), 
      (a_{13}, a_9, a_{15}),\\ 
      (a_{10}, a_{16}, a_{11}),
      (a_2, a_9, a_4), 
      (a_{10}, a_5, a_{11})
      \}
   $ for $i \in \{x, y, z\}$.
   \item $A_x \cup A_y \cup A_z \equiv A_d$ 
     is the multi-agent 
     union set in $F(\Ainit, \nainit)$. 
   \item $\Lambda^{A_d}_{F(\Ainit, \nainit)}
      = \{(a_{13}, a_4, a_{15}), 
      (a_{6}, a_{16}, a_5), 
      (a_{13}, a_9, a_{15}),\\ 
      (a_{10}, a_{16}, a_{11}),
      (a_2, a_9, a_4), 
      (a_{10}, a_5, a_{11})
      \}$.  
\end{itemize} 
}
\noindent $(a_{10}, a_{16}, a_{11})$ and 
$(a_{13}, a_9, a_{15})$, 
as well as $(a_{6}, a_{16}, a_5)$ 
and $(a_{13}, a_4, a_{15})$, 
satisfy (\textbf{Handshake-compatibility}) 
and (\textbf{Resource-safety}).    \\ 
\subsubsection{Haggling and concession.} 
\noindent Let us think of a scenario where $e_2$ 
gets a Switch console from $e_1$ before 
$e_3$ does. The negotiation state $F(\Ainit, \nainit)$ 
transitions to 
the next state $F(A_1, \na_1)$, 
for which we have graphical representation 
of \fbox{B'}, \fbox{C'}, and \fbox{D'} 
from \fbox{B}, \fbox{C}, and \fbox{D}.     
$A_1$ denotes $\{a_{1-2,4-6, 8-11, 12-13,15}\}$ 
with $a_{1,2,\ldots}$ abbreviating $a_1, a_2, \ldots$ 
and $a_{i-j}$ abbreviating $a_i, a_{i+1}, \ldots, 
a_j$. 
\begin{figure}[!h] 
    \includegraphics[scale=0.16]{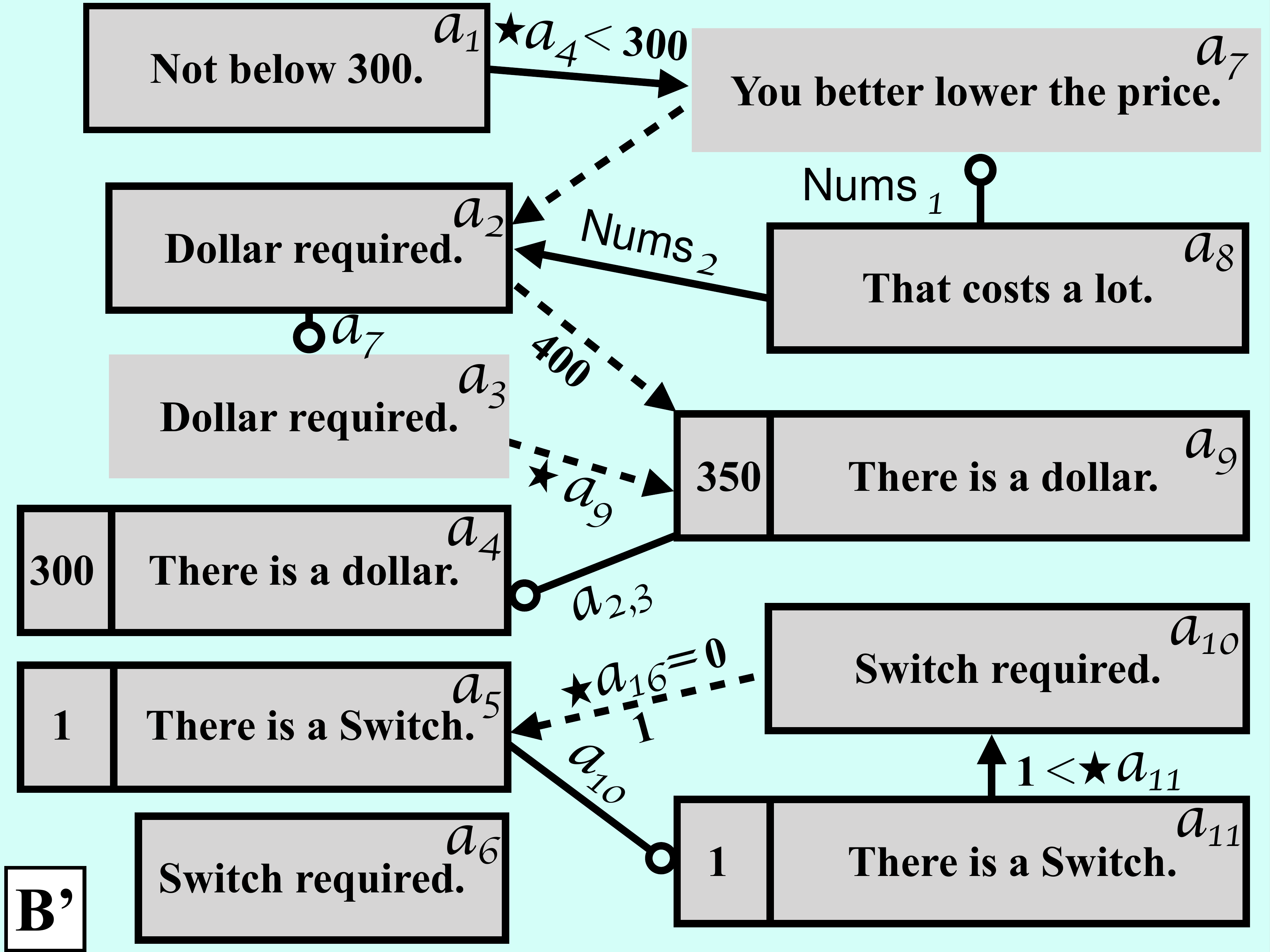} %e2-e3
    \includegraphics[scale=0.16]{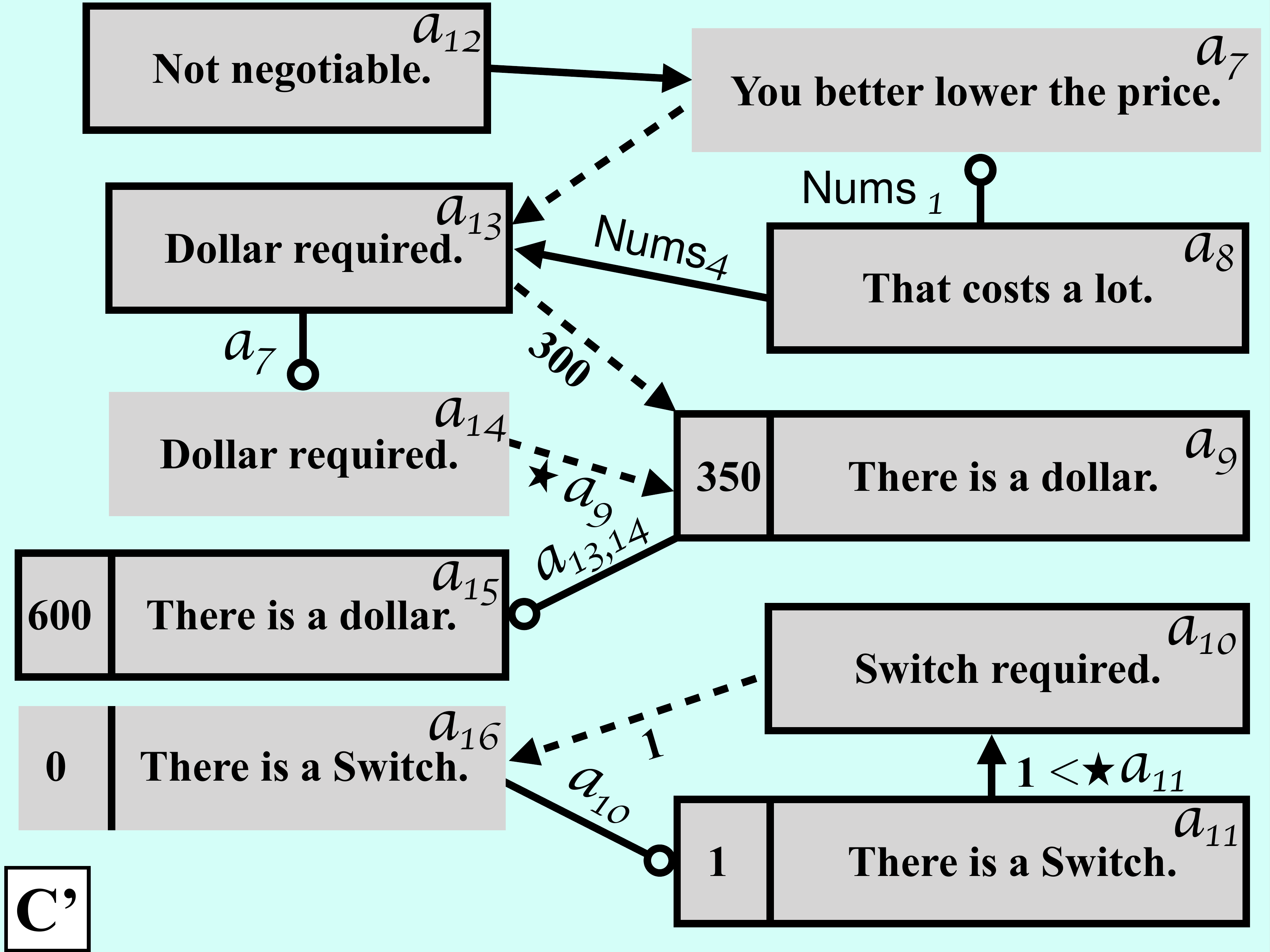} %e1-e2
    \includegraphics[scale=0.16]{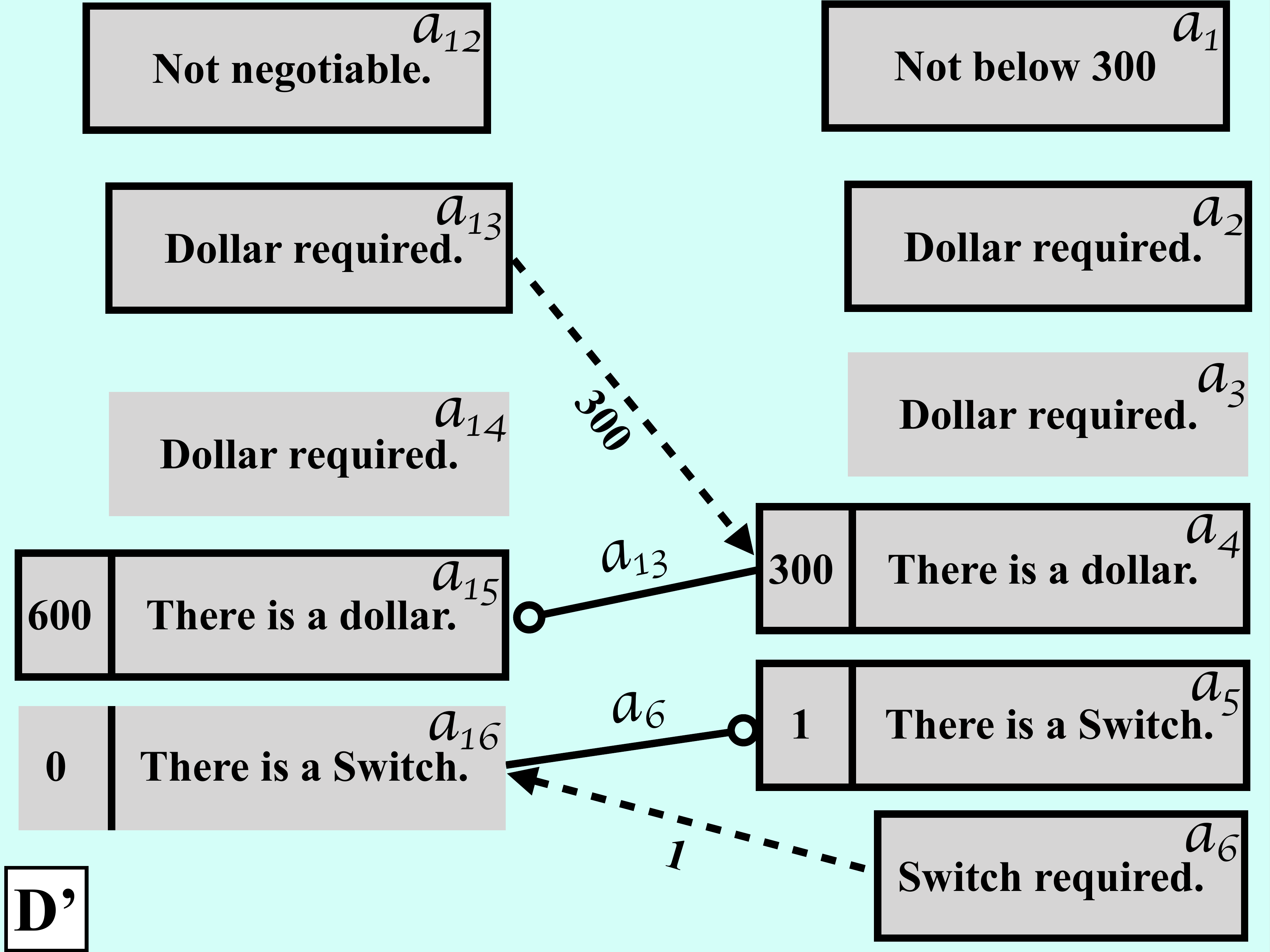} %e1-e3
    \includegraphics[scale=0.16]{blank.pdf} 
\end{figure} 
Meanwhile,  
$\na_1$ is such that 
$\na_1(a_9) = 350$ (reduced from 650 by $e_2$ 
paying 300 to $e_1$), 
that $\na_1(a_{11}) = 1$ (increased from 0 
by $e_2$ obtaining 1 Switch from $e_1$), 
that $\na_1(a_{15}) = 600$ (increased from 300 
by $e_1$ obtaining 300 from $e_2$), 
and that $\na_1(a_{16}) = 0$ (reduced from 1 
by $e_1$ selling 1 Switch to $e_2$), 
while all the other values are unchanged from 
those with $\nainit$. 

At this point, $e_2$ may only obtain  
a Switch from $e_3$. The ask price of a Switch,
however, is 400 dollars from $e_3$, and 
$e_2$ only has 350 dollars. See \fbox{B'} for 
$e_3$-$e_2$ bilateral negotiation. According 
to $e_2$'s preference (find it on the second page), 
$e_2$ asks $e_3$ to lower the ask price to 
350 dollars. That $e_2$ decides to negotiate over the price 
with $e_3$ is characterised in 
$a_8 \overset{\nums_1}{\multimap} 
    a_7 \dashrightarrow 
   a_2 \overset{a_7}{\multimap} 
    a_3$, and by the facts that  
   $\nr((a_3, a_9, a_4)) = \star a_9$, and that
   $\mathcal{I}^t(\star a_9,\na_1) = 350$. 

Since $a_8$ is visible 
   in $F(A_1, \na_1)$, 
   $\sat(\nums_1, \na_1)$ (i.e. $350 < 400$), 
   and no $a \in A_1$ attacks 
   $a_8$ in $F(a_1, \na_1)$, 
   there is a transition from 
   $F(A_1, \na_1)$ into $F(A_2, \na_1)$ 
   with $A_2 = A_1 \cup \{a_7\}$. In this state,
  $e_2$ is trying to convert $e_3$'s $a_2$ with 
   $\nr((a_2, a_9, a_4)) = 400$ 
 into 
   $a_3$ with $\nr((a_3, a_9, a_4)) = \star a_9$ 
   which is basically 350 in $F(A_2, \na_1)$, 
  since $\mathcal{I}^t(\star a_9, \na_1) = 350$. 
  
  Now, the conversion is not automatic
  in all states, because of 
   $a_1 \overset{\star a_4 < 300}{\longrightarrow}{a_7}$. As per $e_3$'s preference 
   (find it on the second page), 
   in case $e_2$ tries to lower the price 
   below 300 dollars, $e_3$ simply refuses  
   the request. In $F(A_2, \na_1)$, however,  
   $\mathcal{I}^t(\star a_4, \na_1) = 350$,
   thus $e_3$ agrees to lower the ask price 
   of a Switch 
   from 400 dollars to 350 dollars, which results in  
   another transition 
   from $F(A_2, \na_1)$ into $F((A_2 \backslash \{a_2\}) 
   \cup \{a_3\}, \na_1)$. 
\section{Conclusion}   
We developed a novel numerical argumentation-based 
negotiation theory from 
a dynamic abstract argumentation \APA, 
and illustrated its mechanism, explaining 
in particular how it deals with preference   
and negotiation process interleaving.  
The following research problems in the literature 
were alleviated or solved by our theory. 
\begin{enumerate} 
  \item Often unexplained origin of preference:   
    compared to the traditional preference that 
    is assumed to be provided from some external source,
    in Numerical {\APA}, whether there are or are not any 
    attacks between arguments is controlled by some numerical values within a given Numerical {\APA} argumentation 
    framework. While we did not explicitly deal with   
    attack-reversing preference \cite{Amgoud14}, 
    with which an attack on a more preferred 
    argument is reversed of its direction, 
    that, too, can be readily handled with numerical 
    conditionalisation. 
  \item Difficulty in handling resource arguments: 
      by quantification of an argument, 
      any potentially diminishing (or strengthening) 
      argument is now intuitively expressed.  
  \item Difficulty in dealing with synchronisation: 
      the handshake mechanism of Numerical {\APA} can be 
      used for this issue.   
  \item Difficulty in handling interleaved negotiations:   
       Numerical {\APA} is an extension of {\APA} 
       which can handle concurrency. Further,
       resource allocation is an inferable, instead of 
       hard-coded, information in Numerical \APA, which
       adapts well to modelling concurrent negotiations. 
\end{enumerate} 
As immediate future work, we have  
two agendas. First, the ability to 
reason about 
game-theoretical properties 
in agent-negotiation is desirable. While 
agents' preferences about the conditions 
under which to permit a resource 
reallocation may be expressed 
by means of  numerical information, 
an inference system will be required 
for judging whether a certain specific 
resource allocation optimal 
to a group of agents is reachable. 
To this end, we consider 
embedding of this theory into ATL \cite{Alur02}, in a similar 
manner to CTL embedding shown 
in \cite{ArisakaSatoh18}. Second,  
we consider relaxation of some assumptions 
made about Numerical \APA.
For instance,  
collaboration among agents may be 
permitted. 
It should be interesting 
to see how different types of agent interactions  
can influence the overall negotiation outcomes. Also, while in this work 
we did not take into account defence from
$k$-step eliminations for 
an arbitrary $k$, any positive 
value for $k$ reflects agent's 
foresight to grasp steps ahead 
in negotiation, which should 
influence its decision-making. 

%Outside argumentation-based negotiations, 
%there are emerging studies, e.g.  
%\cite{Niu18,Zhang17}, on concurrent negotiations. 
%It would be also interesting to seek the role 
%of argumentation in their setting. 
\bibliography{references} 
\bibliographystyle{abbrv}

\end{document}